\documentclass{thesis}
\usepackage[super]{nth}

\author{Tobias Kortkamp}
\matriculationnumber{2491982}

\title{An NLP Assistant for Clide}
\subtitle{}

\date{Monday \nth{26} May, 2014}

\institute{University of Bremen}
\department{Fachbereich 3: Mathematik und Informatik}

\examiner{Prof. Dr. Rolf Drechsler}
\supervisor{Dr. Berthold Hoffmann}
\advisor{Dr. Mathias Soeken and Dipl.-Inf. Martin Ring}

\definecolor{unused-node}{HTML}{828282}
\definecolor{unused-edge}{HTML}{828282}
\definecolor{brackets}{HTML}{b7b7b7}
\definecolor{square-brackets}{HTML}{828282}

\def\clj{Clojure\xspace}
\def\clide{\emph{Clide}\xspace}
\def\clidenlp{\emph{clide-nlp}\xspace}
\def\corenlp{CoreNLP\xspace}
\def\corelogic{\textsf{core.logic}\xspace}

\def\tawnyowl{Tawny-OWL\xspace}

\begin{document}

\maketitle

\tableofcontents

% -*- TeX-master: "../master.tex"; ispell-local-dictionary: "en_US"; -*-
\chapter{Introduction}

%% dass viel natürlichsprachlicher Text im Entwurf von Software und
%% Hardware beschrieben wird, diese Informationen automatisch jedoch
%% kaum genutzt werden können. Deshalb sind immer manuelle Schritte
%% notwendig. Einige einfache Dinge wie Rechtschreibprüfung und
%% Grammatikprüfung gehen automatisch, aber mit allem darüberhinaus
%% ist es schwierig. Es gibt jedoch vereinzelte Ansätze in der
%% Vergangenheit, die versucht haben mehr zu machen, wie z.B.
%% automatische Extraktion in einer formalen Sprache, die dann vom
%% Computer interpretiert werden können. Die Interaktion mit dem
%% Mensch war jedoch oft nicht berücksichtigt worden und somit gab es
%% oft nur rudimentäre Benutzerschnittstellen. Dies Problem kannst du
%% mit clide lösen, da es auf natürliche Weise durch das
%% Kollaborationssystem eine natürliche Schnittstelle zwischen Mensch
%% und Computer herstellt...

While developing software or hardware natural language texts are part
of the process and used to record or specify the system requirements.
The problem is to automatically extract the
information contained in these texts and make them available for
processing. There are some tools that support developers who want to
work with and extract information from these specifications. We are
missing an approachable way to define simple rules to automatically extract and use
information from a text, that is not written in a severely restricted
subset of English.

% While some NLP tools, like e.g. spell checkers or grammar
% checkers, can be run unsupervised, while still yielding acceptable
% results, not all of them can.

Some tools like e.g. Cucumber\footnote{Available at \url{http://cukes.info/}}, a
behavior-driven development framework, solve this by only supporting a
DSL.\footnote{Domain Specific Language} It provides a DSL called Gherkin,
which allows users to write test scenarios that both computers and
humans can understand. Scenarios consist of steps and each step is parsed using 
a user provided regular expression \citep{cucumber}. As a consequence
a step's regular expression is coupled with the specific phrasing that
is used in the step definition. A slight variation in its phrasing
without updating the corresponding regular expression 
or adding a new regular expression, might break the scenario because
the provided regular expression does not match the step anymore.
Ideally, we would like for two steps, with slightly different 
phrasing but the same information content, to yield the same output
and not break the scenario.

Defining and refining these extraction rules is not a solitary
activity, but a collaborative one. By integrating such a system with
\clide,\footnote{Available at \url{https://github.com/martinring/clide2}} a development environment with collaboration baked in, we
support this aspect from the start.

\clide is a project that was developed as part of Martin Ring's
diploma thesis in 2013 at the University of Bremen. It was originally
intended to be a web-based development environment for
Isabelle\footnote{An interactive theorem prover, available at
  \url{http://isabelle.in.tum.de/}} only \citep{ring2013,luth2013}. In
contrast with previous Isabelle interfaces, it provides better
visualization of the prover's results than traditional and sequential
REPL\footnote{Read Eval Print Loop}-based interfaces through
leveraging Web technologies like HTML5 and JavaScript
\citep{ring2013}. It has since undergone further development and has
evolved to facilitate collaborative editing of documents with support
for other languages besides Isabelle \citep{ring2014}.

\clide documents are annotated by assistants. It uses an approach
called \emph{Universal Collaboration} where an assistant is seen as
just an additional collaborator by the system \citep{ring2014}. While
\clide is distributive and asynchronous in nature, it provides an
interface that can be used to implement assistants ``in a simple,
synchronous manner'' \citep{ring2014}.

A \clide assistant is informed about state changes of \clide's
environment, be it newly opened documents or edits of a document. It can
provide domain specific annotations for any parts of a document's text. 

\begin{figure}
  \includegraphics[width=\textwidth]{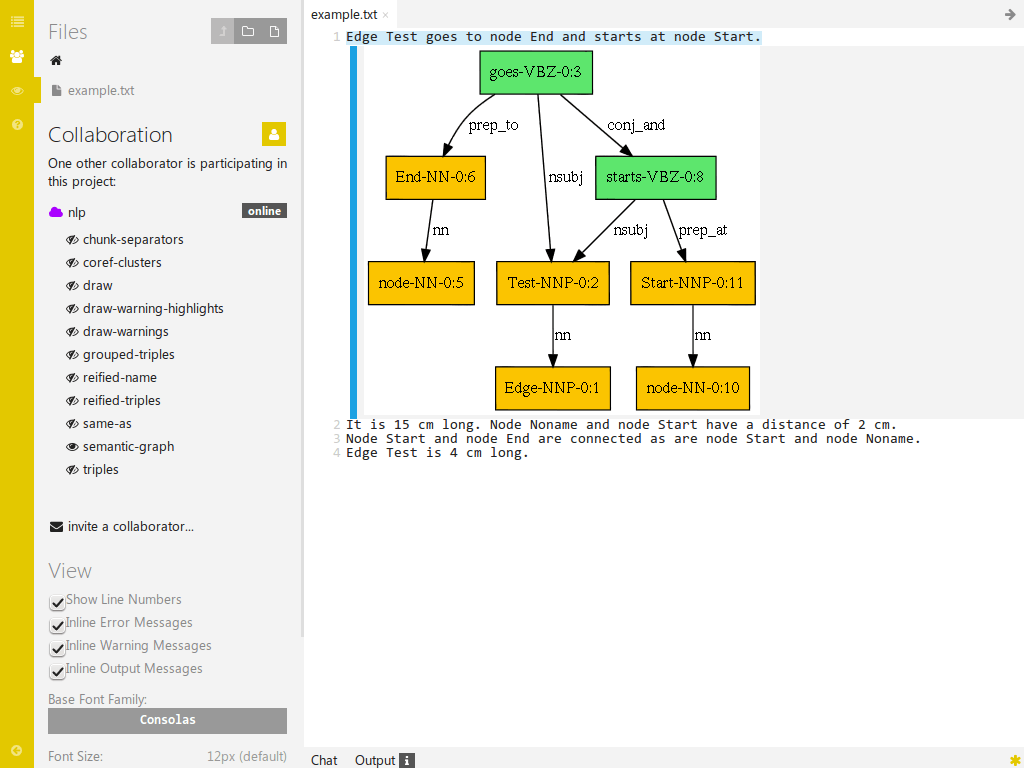}
  \caption{The \clidenlp annotation \textsf{semantic-graph}
    showing the semantic graph for the current sentence (highlighted in blue).}
  \label{fig:screen-semantic-graph}
\end{figure}

\begin{figure}
  \includegraphics[width=\textwidth]{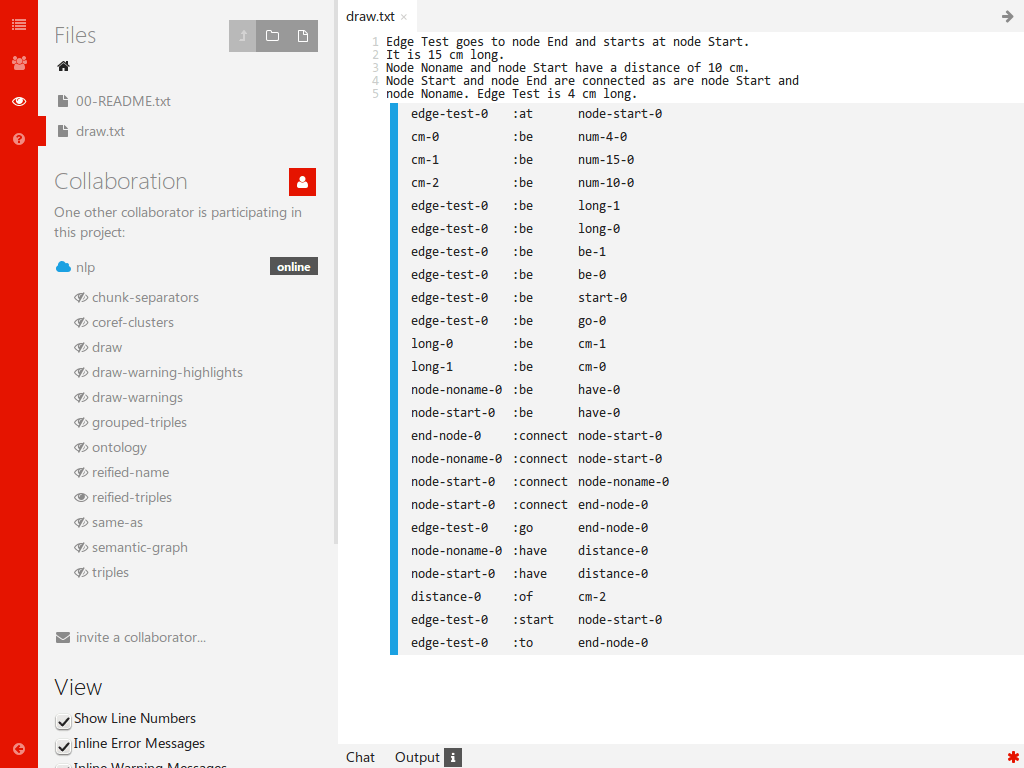}
  \caption{The \clidenlp annotation \textsf{reified-triples}
    showing the triples that are found for the given text and the
    annotation \textsf{same-as} highlighting all words that are
    detected as belonging together in red. The highlighted words are
    part of the group \textsf{Edge-Test-it-0}.}
  \label{fig:screen-reified-triples}
\end{figure}

This report describes \clidenlp, an NLP\footnote{Natural Language
  Processing} assistant for \clide. The assistant has the following
goals:
\begin{itemize}
\item Create a framework for extracting ontologies from a text, by
  \begin{itemize}
    \item creating an NLP knowledge base (see
      \prettyref{cha:approach}), and
    \item using simple queries on that knowledge base to extract
     useful information from the text (see \prettyref{cha:triples}).
  \end{itemize}
\item Provide annotations for interacting with a text to assist in
  developing of the queries, including showing
  \begin{itemize}
  \item the semantic graph of a sentence (see \prettyref{fig:screen-semantic-graph}),
  \item the coreferences\footnote{Groups of related entities in a text} of the text, and
  \item the ontology extracted from the text (see \prettyref{fig:screen-reified-triples}).
  \end{itemize}
\item Work in the collaborative environment that \clide provides and
  keep the ontology and annotations up to date and in sync with text
  changes. 
\item Being one of the first \clide assistants not developed by
  \clide's author, \clidenlp should also be
  seen as a test of the limits of \clide's current API for developing
  assistants. \clide's development continues in parallel with the work
  on this report.
\end{itemize}

\prettyref{cha:approach} introduces the components that constitute \clidenlp and
how they interact with each other. \prettyref{sec:architecture} and
\prettyref{sec:clide} discuss how \clidenlp is integrated into \clide.

Because of \clide's interactive and collaborative nature, \clidenlp
has to contend with continuously changing text.
\prettyref{sec:reconciler} discusses a model for providing support for
incorporating text changes into \clidenlp's internal model.
The provided annotations need to reflect these changes immediately, or
as fast as possible.

\clidenlp uses the \corenlp framework by the Stanford University's NLP
group that provides access to several NLP tools, including a
coreference resolution system, which groups related entities in a text together,
and a dependency parser, that describes the relations between sentence
parts. \prettyref{sec:corenlp} and \prettyref{sec:knowledge-base}
describe how we can build an NLP knowledge base based on these tools.

\prettyref{cha:triples} shows how we can leverage the knowledge base
to extract an ontology through a series of simple queries. The queries
are described in detail in \prettyref{sec:triple-builders}.
\prettyref{sec:reified-triples} describes how the queries' results are
used to create an ontology. \prettyref{sec:ontology} discusses one possible
way of how the ontology can be exported to an OWL\footnote{Web
  Ontology Language} ontology.

\begin{figure}
  \includegraphics[width=\textwidth]{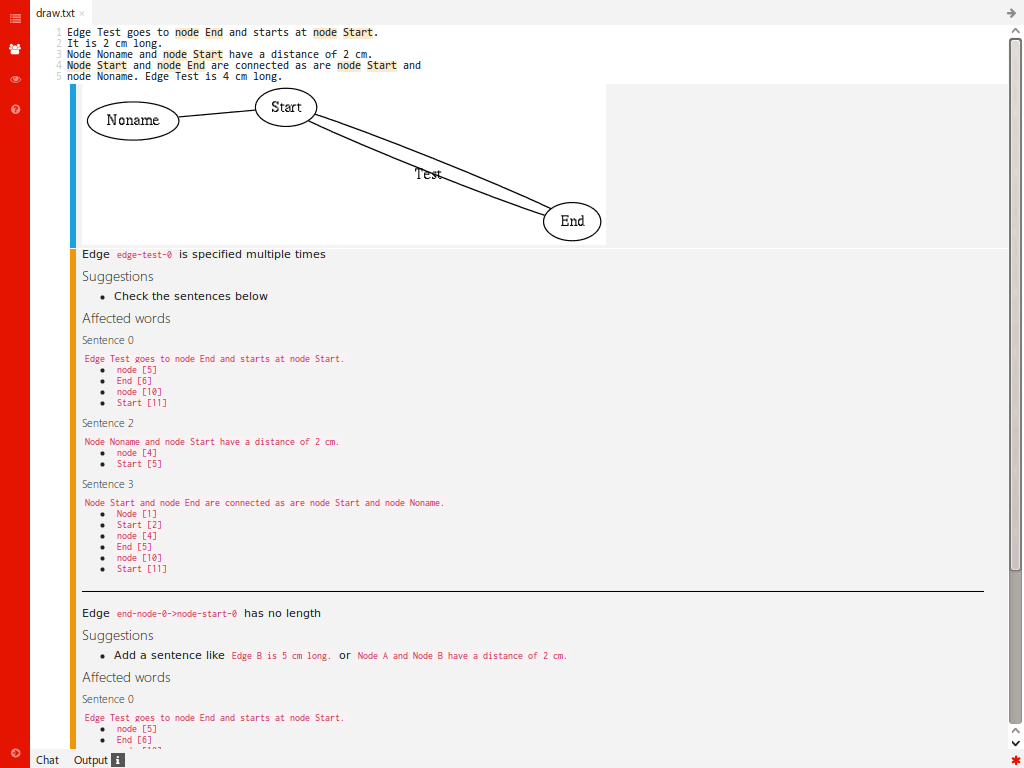}
  \caption[Showing all annotation from the example application
  \textsf{draw}, \textsf{draw-warnings} and
  \textsf{draw-warning-highlights} that draws the graph
  specified in the text.]{Showing all annotation from the example
    application \textsf{draw}, \textsf{draw-warnings} and
    \textsf{draw-warning-highlights} that draws the graph specified in
    the text and highlights potential problems and ambiguous values in
    the text.}
  \label{fig:screen-draw}
\end{figure}

In \prettyref{cha:usecase} we describe an example application that uses the
extracted ontology. We extend \clidenlp to create graphs from
natural language specifications by leveraging the ontology.
\prettyref{fig:screen-draw} shows an example graph.

% -*- TeX-master: "../master.tex"; ispell-local-dictionary: "en_US"; -*-

\chapter{Basics}

This chapter introduces some of the concepts needed to
understand this report.

\clidenlp is implemented in \clj. We first give a short overview of
\clj and its logic programming library \corelogic. We conclude this
chapter with a short introduction of \corenlp, the NLP framework used
by \clidenlp.

\section{\clj}

\clj is a functional programming language and has
several features that make it a good fit for NLP applications.

\clj provides a REPL which allows us to patch in new code into a
running system, enabling a small ``thought-code-feedback loop''
\citep{fogus2011}.

It provides built-in persistent data structures with their own reader
syntax, that are immutable and have clear and intuitive value equality
semantics (1 is 1 and \lstinline|[1 2 3]| is \lstinline|[1 2 3]|).

\clj is built on top of the JVM and has good interoperability support with
Java libraries. This allows us to leverage all existing Java NLP
libraries \citep{cemerick2012,ling-clojure}.

\clj's macros can be used to extend the language when needed. This is
used heavily by \corelogic, which adds logic programming to \clj.

The following table summarizes the aspects of \clj's syntax that is
important for reading this report:

\begin{tabular}{l|l|p{9cm}}
  \bf Type & \bf Example & \bf Description \\
  \hline
  Function definition & \lstinline|(defn $\textsf{f}$ [$x$ $y$] $\dots$)|
  & \textsf{defn} defines a function. Here we define the function
  \textsf{f} that takes 2 arguments $x$ and $y$.\\
  Function call & ($\textsf{f}$ $x$) & \clj is a Lisp and uses prefix
  notation. Here we call the function \textsf{f} with argument $x$. \\
  Keyword & \textsf{:a} & Keywords are often used as keys in a map, because
  they evaluate to themselves and can be used as functions that look
  themselves up in the associative data structures (e.g. a map) that
  is passed to them as their first argument.\\
  Map & \lstinline|{$\textsf{:a}$ "Hello" $\textsf{:b}$ 0}| & A map with the
  key-value pairs (\textsf{:a}, \lstinline|"Hello"|) and (\textsf{:b},
  0)\\
  Vector & \lstinline|[1 2 "Hello"]| & A vector with the elements 1,
  2 and \lstinline|"Hello"|\\
  \hline\hline
\end{tabular}

There is no special interoperability support for interfacing
with Scala libraries, like e.g. \clide. Interfacing with Scala
directly from Java is already challenging, and interfacing with Scala
from \clj adds an additional complication. The work around is to
write the components that use \clide directly in Scala and use
\clj's runtime interface to call into \clj code from Scala.

More information on \clj is available in \citep{cemerick2012,fogus2011}.

\subsection{Logic programming with \corelogic}
\label{sec:corelogic}

\def\minikanren{miniKanren\xspace}
\def\ckanren{cKanren\xspace}

\corelogic\footnote{Available at \url{https://github.com/clojure/core.logic}}
adds logic programming capabilities to \clj. It is based
on \minikanren, a logic programming library for Scheme, developed by
William E. Byrd as part of his PhD thesis \citep{minikanren}.

Because \corelogic is a library for \clj, we can mix functional and
logic programming freely, dropping down to \corelogic when we need it
and use \clj functional programming aspects otherwise \citep{ling-clojure}.

% Rather than introducing \clj in more detail, we focus on introducing
% \corelogic through a series of simple examples, to allow the reader to
% develop some intuition for reading \corelogic based code.

We summarize the most important functions and macros of \corelogic
here:

\begin{tabular}{l|p{4cm}|p{7cm}}
  \bf Type & \bf Example & \bf Description\\
  \hline
  Run a query & \lstinline|(run*$\ $[$q$] $\dots$)| & Runs a \corelogic
  query by trying to unify a value with $q$. Returns a list of all
  possible values for $q$\\

  Create logic variables & \lstinline|(fresh$\ $[$a$ $b$] $\dots$)| & Creates two unbound logic variables $a$ and $b$\\

  Unnamed logic variable & \lstinline|(lvar)| & Returns a new logic variable\\

  Logical disjunction & {
    \begin{lstlisting}[style=inline,gobble=6,belowskip=-1.5em,aboveskip=-1em]
      (conde
        [<$\textrm{branch1}$>]
        [<$\textrm{branch2}$>]
        $\dots$)\end{lstlisting}
  } & Tries all branches consecutively \\

  Soft cut & \lstinline|(conda$\ \dots$)| & Like \lstinline|conde| but
  stops the search as soon as a branch succeeds\\

  Feature extraction & {
    \begin{lstlisting}[style=inline,gobble=6,belowskip=-1.5em,aboveskip=-1em]
      (featurec
        {$\textsf{:a}$ 4 $\textsf{:b}$ 5}
        {$\textsf{:a}$ $q$})\end{lstlisting}
  } &
  Extracts features from maps. The example binds the logic variable
  $q$ to 4.\\
  Unify & \lstinline|(== $q$ 4)| & Unifies the logic variable $q$ with 4.\\
  Never unify & \lstinline|(!= $q$ 4)| & Adds a constraint to the
  logic variable $q$ that it can never be 4.\\
  List membership & \lstinline|(membero $q$ [1 2 3])| & A goal that succeeds if $q$ is bound to value that is in the vector \lstinline|[1 2 3]|\\
  \multirow{1}{4cm}{Extract a logic variable's value inside a query} & \lstinline|(project$\ $[$q$]$\ \dots$)| & Extracts the value that is bound to the logic variable $q$. While in scope of a \textsf{project} $q$ is a regular Clojure value and we can use regular Clojure functions with it.\\
  Domain constraint & \lstinline|(in $q$ (interval 1 10))| & Makes sure that $q$ is bound to a value in the interval $[1, 10]$.\\
  \hline\hline
\end{tabular}

\newpage
The presentation of the \corelogic code shown here is based on the
code presentation in \citep{reasoned-schemer,minikanren}.

\begin{tabular}{l|l|p{9.5cm}}
  \bf Presentation & \bf Actual code & \\
  \hline
  \lstinline|<x>$^\circ$| & \texttt{<x>o} &
  A goal is written with a suffix \textsf{o} to distinguish it from already defined functions on the functional programming side, while making clear that they have the same outcome in both paradigms \citep{minikanren}. E.g. \textsf{cons$^\circ$} in \corelogic and \textsf{cons} in \clj\\
  \lstinline|run*| & \texttt{run*}\\
  \lstinline|conde| & \texttt{conde} & \multirow{2}{9cm}{The branching macros have an added suffix to distinguish them from the built-in \textsf{cond}.} \\
  \lstinline|conda| & \texttt{conda} & \\
  \lstinline|==| & \texttt{==} & \\
  \lstinline|!=| & \texttt{!=} & \\
  \hline\hline
\end{tabular}

\emph{The Reasoned Schemer} \citep{reasoned-schemer} provides a good
introduction to \minikanren and in extension also \corelogic.\footnote{Their differences are described on \corelogic's
  Wiki available at \url{https://github.com/clojure/core.logic/wiki/Differences-from-The-Reasoned-Schemer}.}

\begin{example} We create a new knowledge base and populate it with
  three facts about animals.

  \begin{center}
  \begin{tabular}{c|c|c}
      \bf Relation & \bf Kind & \bf Name \\
      \hline
      \sf animal & cat & Felix \\
      \sf animal & cat & Mittens \\
      \sf animal & dog & Waldo \\
      \hline\hline
    \end{tabular}
  \end{center}
  \vspace*{-1em}
  \end{example}

Using this knowledge base, we can define a new goal that succeeds iff
an animal is a cat or a dog. We make use of the predefined goal
\lstinline|membero|to check if the value of a logic variable is inside
of a collection.
\begin{lstlisting}
(defn $\textsf{cat-or-dog}^\textsf{o}$ [$\textrm{name}$ q]
  (fresh [t]
    (animal t $\textrm{name}$)
    (membero t ["cat" "dog"])
    (== t q)))
\end{lstlisting}

We can then build a query to check what kind of an animal Waldo is:
\begin{lstlisting}
(run* [q]
  ($\textsf{cat-or-dog}^\textsf{o}$ "Waldo" q))
$\Rightarrow$ ("dog")
\end{lstlisting}

Waldo is a dog! And Benjamin?

\begin{lstlisting}
(run* [q]
  ($\textsf{cat-or-dog}^\textsf{o}$ "Benjamin" q))
$\Rightarrow$ ()
\end{lstlisting}

Benjamin does not exist in the knowledge base, so the query returns no result.

\begin{minipage}{\textwidth}
We can run the query in reverse to get all cats:
\begin{lstlisting}
(run* [q]
  ($\textsf{cat-or-dog}^\textsf{o}$ q "cat"))
$\Rightarrow$ ("Felix" "Mittens")
\end{lstlisting}
\end{minipage}

While \corelogic supports relational programming, our usage of several
non-relation goals, like \lstinline|conda| or \lstinline|project|,
makes all of our \corelogic usage effectively non-relational.

\subsection{\tawnyowl}

\tawnyowl\footnote{Available at
  \url{https://github.com/phillord/tawny-owl}} is a \clj library that
provides a domain specific language for building OWL ontologies
\citep{lord2013}.

\clidenlp uses \tawnyowl for building OWL ontologies out of its custom
ontologies it extracts from texts (see \prettyref{sec:ontology}).
Exporting OWL ontologies allows us to make use of the existing OWL
tools, like e.g. querying them via SparQL \citep{semantic-web}.

\section{\corenlp}

\corenlp is an NLP framework that was created by the NLP group at
Stanford University.\footnote{Available at
  \url{http://nlp.stanford.edu/}} It includes several components that
facilitate developing NLP applications or algorithms. \clidenlp uses
\corenlp's dependency parser and its coreference resolution system.

The dependency parser makes the underlying structures of
sentences visible in the form of grammatical relations between
sentence parts \citep{de2006}. The output of this component can be
modeled as a graph, where the grammatical relations are the
graph's edges and the nodes are the sentence parts. We call this
graph \emph{semantic graph} in this report. Examples of semantic
graphs are available in \prettyref{sec:triple-builders} and
\prettyref{sec:corenlp}. The grammatical relations are described
in \citep{dependencies-manual}. The dependency parser can collapse
prepositions and coordinations into grammatical relations
\citep{de2006,dependencies-manual}. \clidenlp uses this feature to
simplify the resulting semantic graphs.

The dependency parser makes use of \corenlp's part-of-speech
(POS) tagger. Its tagset is based on the POS tagset used by the
Penn Treebank,  described in \citep{marcus1993}.
Because semantic graphs contain POS tags and \clidenlp makes heavy
use of semantic graphs, \prettyref{tab:pos}
provides an overview over some of the tags provided by \corenlp.

The deterministic coreference resolution system is used to
identify entities refer to each other (also called mentions).
It was introduced in
\citep{raghunathan2010,lee2011,lee2013,recasens2013}. It competed
in the CoNLL Shared Task 2011, where it achieved the highest score
in both the closed and open tracks \citep{lee2011}.

\prettyref{sec:coreference-cluster-mentions} goes into more detail and
shows an example coreference cluster.

% \input{parts/executive_summary.tex}

% -*- TeX-master: "../master.tex"; ispell-local-dictionary: "en_US"; -*-

\chapter{Approach}
\label{cha:approach}

This chapter introduces the components that constitute \clidenlp and
how they interact with each other. We first describe how data flows
between the components by giving a high-level architecture overview in
\prettyref{sec:architecture}. We then delve deeper into the implementation.

\clidenlp receives a continuous stream of events (text operations, cursor
movements, and annotation requests) from \clide that need to be integrated
with \clidenlp's underlying computation model.
\prettyref{sec:reconciler} describes this model and how we integrate
changes into it.

In \prettyref{sec:corenlp} we continue with a description of how
\corenlp is integrated into the system by revealing some pitfalls that
occur when using the diverse data structures provided by \corenlp and
how we can avoid them. In \prettyref{sec:knowledge-base} we then build a
knowledge base from the data provided by \corenlp which we can use with
\corelogic and that forms the basis for the remaining chapters.

We conclude this chapter with \prettyref{sec:clide} and explain the
user-facing side of \clidenlp by showing what annotations are provided
and what caveats apply given the current annotation model of \clide.

\section{Architecture}
\label{sec:architecture}

\def\archrelation#1{%
{$\vcenter{\hbox{\scalefont{0.75}%
\begin{tikzpicture}[>=latex]\node[draw,thick] (a) {$A$};
  \node[right=7mm of a,draw,thick] (b) {$B$} edge [draw,thick,#1] (a);\end{tikzpicture}}}$}}

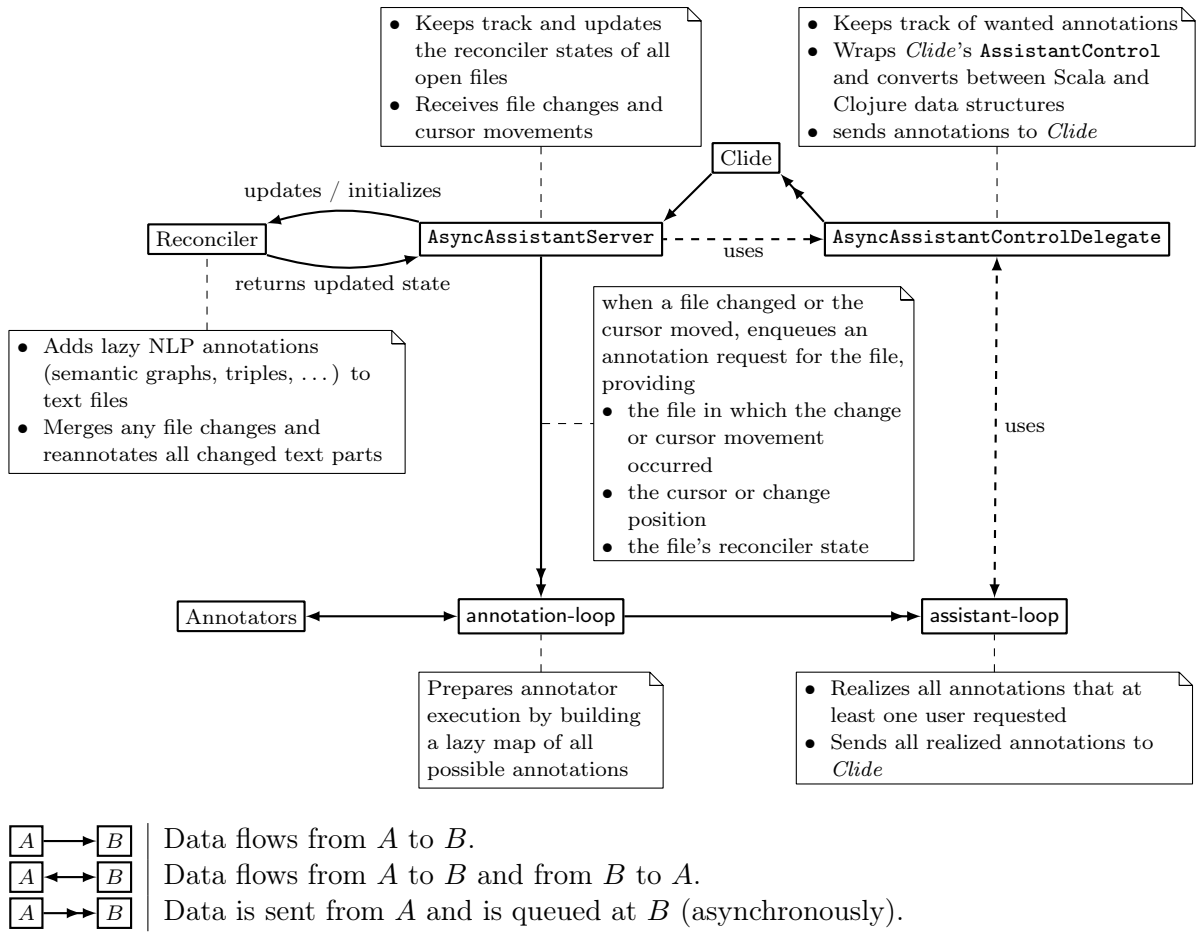
\begin{figure}[h!]
  {\scalefont{0.75} \begin{center}
\begin{tikzpicture}[>=latex,line join=bevel,thick,node distance=2cm]

  \node (assistant_server) [draw,rectangle] {\texttt{AsyncAssistantServer}};
  
  \node (clide) [above right=0.9cm of assistant_server,draw,rectangle] {Clide};

  \node (assistant_control) [right of=assistant_server,node distance=6cm,draw,rectangle] {\texttt{AsyncAssistantControlDelegate}}
  edge[<-,dashed] node[below] {uses} (assistant_server);

  \path[] (assistant_server.north east) edge[<-] (clide.south west);
  \path[] (assistant_control.north west) edge[->>] (clide.south east);
  
  \node (annotation_loop) [below of=assistant_server,node distance=5cm,draw,rectangle] {\textsf{annotation-loop}};

  \node (annotation_loop_note1) [xshift=2.8cm,yshift=-1mm,below=1em of assistant_server,note,thin,draw] {\begin{minipage}[l]{4cm}
      \begin{flushleft}
        when a file changed or the cursor moved, enqueues an annotation request
        for the file, providing
        \begin{compactitem}[$\bullet$]         
        \item the file in which the change or cursor movement occurred
        \item the cursor or change position
        \item the file's reconciler state
        \end{compactitem}
        \end{flushleft}
      \end{minipage}};    
    \node (annotation_loop_note1_edge) [below=of assistant_server,yshift=-1mm,xshift=-1mm] {}
    edge [-,dashed,thin] (annotation_loop_note1);
    \path (annotation_loop) edge[<<-] (assistant_server);
  
  \node (assistant_loop) [right=5cm of annotation_loop.center,draw,rectangle] {\textsf{assistant-loop}}
  edge[<<-] node[below] {} (annotation_loop)
  edge[<->,dashed] node[right] {uses} (assistant_control);
  
  \node (annotators) [left=2cm of annotation_loop,draw,rectangle] {Annotators}
  edge[<->] node[below] {} (annotation_loop);

  % \node(annotators_note) [below=0.5cm of annotators,draw,note,thin] {
  %   \begin{minipage}[c]{3cm}
  %     \begin{flushleft}
  % % TODO
  %       NLP annotations are only realized when an Annotator needs them
  %     \end{flushleft}
  %   \end{minipage}}
  % edge[-,dashed,thin] (annotators);
  
  \node (reconciler) [left=2cm of assistant_server,draw,rectangle] {Reconciler};
  \path[] (reconciler.north east) edge[<-,bend left=15] node[above] {updates / initializes} (assistant_server.north west);
  \path[] (reconciler.south east) edge[->,bend right=15] node[below] {returns updated state} (assistant_server.south west);

  \node (reconciler_note) [below=1cm of reconciler,draw,note,thin] {\begin{minipage}[c]{5cm}
      \begin{flushleft}
      \begin{compactitem}[$\bullet$]
      \item Adds lazy NLP annotations (semantic graphs, triples, \dots) to text files
      \item Merges any file changes and reannotates all changed text parts
      \end{compactitem}
      \end{flushleft}
    \end{minipage}}
  edge[-,dashed,thin] (reconciler);  
  
  \node (assistant_control_note) [above=1cm of assistant_control,draw,note,thin] {\begin{minipage}[c]{5cm}
      \begin{flushleft}
      \begin{compactitem}[$\bullet$]
      \item Keeps track of wanted annotations
      \item Wraps \clide's \texttt{AssistantControl} and converts between Scala
        and Clojure data structures
        \item sends annotations to \clide 
      \end{compactitem}
      \end{flushleft}
    \end{minipage}}
  edge[-,dashed,thin] (assistant_control);
  \node (assistant_server_note1) [above=1cm of assistant_server,draw,note,thin] {\begin{minipage}[c]{4cm}
      \begin{flushleft}
      \begin{compactitem}[$\bullet$]
      \item Keeps track and updates the reconciler states of all open files        
      \item Receives file changes and cursor movements       
      \end{compactitem}
      \end{flushleft}
    \end{minipage}}
  edge[-,dashed,thin] (assistant_server);

  \node (annotation_loop_note) [below=0.5cm of annotation_loop,draw,note,thin] {\begin{minipage}[c]{3cm}
      \begin{flushleft}
        Prepares annotator execution by building a lazy map of all possible annotations
      \end{flushleft}
    \end{minipage}}
  edge[-,dashed,thin] (annotation_loop);

  \node (assistant_loop_note) [below=0.5cm of assistant_loop,draw,note,thin] {\begin{minipage}[c]{5cm}
      \begin{flushleft}
      \begin{compactitem}[$\bullet$]
      \item Realizes all annotations that at least one user requested
      \item Sends all realized annotations to \clide
      \end{compactitem}
      \end{flushleft}
    \end{minipage}}
  edge[-,dashed,thin] (assistant_loop);
\end{tikzpicture}
\end{center}}
  \begin{tabular}{l|l}
    \archrelation{<-} & Data flows from $A$ to $B$.\\  
    \archrelation{<->} & Data flows from $A$ to $B$ and from $B$ to $A$.\\    
    \archrelation{<<-} & Data is sent from $A$ and is queued at $B$ (asynchronously).\\  
  % \archrelation{<-,dashed}  & $A$ uses functionality from $B$\\
  % \archrelation{<->,dashed} & $A$ uses $B$ and $B$ uses $A$\\
  % \archrelation{-,dashed} & Note
  \end{tabular}
  \caption[Architecture overview]{High-level architecture overview showing the data flow
    between \clidenlp's components}
  \label{fig:architecture}
\end{figure}

\prettyref{fig:architecture} shows how data flows between the
components in \clidenlp.

\clide assistants need to provide a subclass of
\texttt{AssistantServer}. \texttt{AssistantServer} has support for
connecting and receiving messages from \clide built-in and abstracts away
the underlying Akka implementation.

\clidenlp calls its \texttt{AssistantServer} subclass
\texttt{AsyncAssistantServer}. It receives events for file changes and
cursor movements.

All file changes are passed to the reconciler, which uses
referentially transparent functions. We pass them the current state of
a file and it returns an updated version of that state. The
\texttt{AsyncAssistantServer} instance is responsible for storing that 
state and retrieving it when needed.

The \texttt{AsyncAssistantServer} enqueues the reconciler state and the
cursor position of the change in a queue that is used by the
\textsf{annotation-loop}.

The \textsf{annotation-loop} reads one element (state and position) at
a time from the queue and prepares \clide annotations based on it. See
\prettyref{sec:clidenlp-annotations} for a list of provided
annotations. 

The annotations are not computed in the \textsf{annotation-loop}, but
are sent to the \textsf{assistant-loop}, which is responsible for
realizing only those annotations that the users of \clidenlp want to see and
then sending them to \clide.

The annotators that create \clide annotations make use of the
reconciler's lazy NLP annotations (see \prettyref{sec:reconciler}).
The combination of lazy \clide annotations and lazy NLP annotations
guarantees that \clidenlp only does work when it really has to.

% -*- TeX-master: "../master.tex"; ispell-local-dictionary: "en_US"; -*-

\section{Reconciler}
\label{sec:reconciler}

\clide is a collaborative editing environment. Multiple users may
change the current file's text at any time. As such \clidenlp needs a
way to incorporate those text changes into its own data model.

In traditional IDEs the process which incorporates changes into its
data models is called \emph{reconciliation}. \clide itself does not
provide built-in support for reconciling yet. \clidenlp has to provide
that support itself.

\begin{minipage}{\textwidth}
The reconciler has several related tasks, which are performed in the
following order:
\begin{enumerate}
  \item Split the input text into separate chunks
  \item Replay \clide deltas to incorporate text changes and mark all
    chunks that have changed
  \item Compute NLP annotations for each changed chunk
\end{enumerate}
\end{minipage}

\subsection{Chunking}

\clidenlp splits an input text into several parts to keep the time
needed for (re-)computing the NLP annotations to a minimum and to make
testing easier.

In the implementation provided with this report, \clidenlp splits an
input text at the string \verb|"\n----\n"|.

\begin{example}
  The following text
  
\begin{verbatim}
The cat eats the mouse.

----

The mouse is dead.
\end{verbatim}

is split into 3 chunks:
\begin{enumerate}
  \item The cat eats the mouse.
  \item \verb|----|
  \item The mouse is dead.
\end{enumerate}
\end{example}

Each chunk has an associated span. A span is the offset interval from the
beginning of the text. In the example above, Chunk 1 has a span of
$[0,25)$ and Chunk 2 is called a chunk separator.

Because a chunk is just a slice of an input text, more complicated
chunkers are possible, and indeed would be more useful and realistic
than the very simplistic chunker currently implemented.

We could e.g. treat comment blocks in a Java
file as a chunk and the code in between blocks as chunk separators.
\corenlp might not understand tokens or characters used in Java
comments and as a consequence, we would need to remove them first.
If we replace them with whitespace before passing the comment to
\corenlp, we make sure that we can map to the original text in an easy
way by mirroring the spans inside the original text and inside the
replacement text.\footnote{\corenlp does something similar in its
  \texttt{cleanxml} annotator to remove XML tags from an input text
  before parsing it.}

\begin{example}
The Java comment
\begin{verbatim}
/**
 * This is a comment
 */
\end{verbatim}
can be written as $$S = \textsf{\lstinline|"/**$\backslash n$ * This is a comment$\backslash n$ */"|}$$
% \texttt{``/**$\backslash n$\_*\_This\_is\_a\_comment$\backslash n$\_*/''}$$ with spaces
The substring $$T = \textsf{\lstinline|"This is a comment"|}$$ has the span
$[7, 24)$ in $S$.
If we replace all special characters not understood by \corenlp with
spaces, we would get the string
$$S' = \textsf{\lstinline|"$\textsf{\textvisiblespace\textvisiblespace\textvisiblespace}\backslash n$   This is a comment$\backslash n\textsf{\textvisiblespace\textvisiblespace\textvisiblespace}$"|}$$
The span of $T$ in $S'$ is still $[7, 24)$.
\end{example}

\subsection{Incorporating text changes}
\label{sec:reconciler-text-changes}

In \clide text changes are described as a list of operations that
describe the steps needed to transform an old version of a text
into a new version. There are three operations \citep{ring2014}:
\begin{itemize}
\item \texttt{Retain($n$)}
\item \texttt{Insert($s$)}
\item \texttt{Delete($n$)}  
\end{itemize}

Since \clide is written in Scala \texttt{Retain}, \texttt{Insert} and
\texttt{Delete} are implemented using Scala case classes.\footnote{Case
  classes are algebraic data type constructors and allow pattern matching.} While it is
possible to work with theses classes in Clojure, it is easier to
translate them to use Clojure's data structures instead. 

The translation is straightforward. For each operation, replace

\begin{tabular}{rcl}
    \texttt{Retain($n$)} & with & \textsf{[:retain $n$]}\\
    \texttt{Insert($s$)} & with & \textsf{[:insert $s$]}\\
    \texttt{Delete($n$)} & with & \textsf{[:delete $n$]}
\end{tabular}

\begin{example}
The operations \texttt{[Retain(5),
  Insert(\lstinline|"hallo"|), Retain(15)]} are translated into the Clojure data
\textsf{[[:retain 5] [:insert \lstinline|"hallo"|] [:retain 15]]}.
\end{example}

To apply the operations we need to maintain a cursor position starting
at 0. The cursor's position is the position in the original text, not
the edited text. The original text is immutable and is not changed.
The edited text is built by applying each operation sequentially:
\begin{itemize}
\item \textsf{[:retain $n$]} moves the cursor $n$ characters ahead and
  inserts them in the edited text.
\item \textsf{[:insert $s$]} inserts the string $s$ at the current
  cursor position in the edited text. This will not move the cursor,
  because it is relative to the original text.
\item \textsf{[:delete $n$]} deletes the next $n$ characters at the
  current cursor position and moves the cursor $n$ characters ahead.
\end{itemize}

\begin{example}
Given the input text \lstinline|"This is a test."| and the operations
\textsf{[[:retain 9] [:insert \lstinline|"the"|] [:delete 1] [:retain 6]]}, we
get the text \lstinline|"This is the test."| after applying them.
\end{example}

As the reconciler splits a text into several chunks there are some
more concerns to address.

The changes to a text can be
\begin{enumerate}
\item inside a chunk separator, i.e. between two chunks $A$ and
  $B$\footnote{Chunk $B$ follows chunk $A$.} in which case there are two possible
  scenarios:
  \begin{enumerate}[a.]
  \item $A$ remains unchanged and the spans of $B$ and of all chunks
    that follow it have to be updated.
  \item $A$ and $B$ have to be merged together because the chunk
    separator is not valid anymore. This would also change the
    spans and indices of all chunks that follow $A$ and $B$.
  \end{enumerate}
\item inside one chunk $C$, in which case $C$'s span (end offset) and
  the span of all chunks that follow $C$ must be updated to
  reflect the changes.
\end{enumerate}

In the actual reconciler implementation changes between two chunks % (in a chunk separator)
are detected after applying the edit operations and rechunking the text.
If the number of chunks changed, the reconciler is simply
reinitialized. This greatly simplifies the implementation, but all NLP
annotations are lost and need to be rebuilt. An ideal implementation
would have to follow all scenarios above.

\subsection{Chunk annotations}

Each chunk has associated annotations that are updated when a
chunk changes. The annotations are added to a Clojure map with
lazy evaluation semantics. This model allows the reconciler to remain
fast even when there are continuous changes. The chunk annotations are only
realized, and thus computed, outside of the reconciler.

\begin{figure}[h!]
  \centering
  \begin{tikzpicture}[>=latex,line join=bevel,]
\begin{scope}[thick,inner sep=4]
  \node (_corenlp-annotation) at (136bp,318bp) [draw,draw=none] {\textsf{:corenlp-annotation}};
  \node (_grouped-triples) at (88bp,162bp) [draw,draw=none] {\textsf{:grouped-triples}};
  \node (_semantic-graphs) at (136bp,266bp) [draw,draw=none] {\textsf{:semantic-graphs}};
  \node (_reified-triples-knowledge-base) at (152bp,58bp) [draw,draw=none] {\textsf{:reified-triples-knowledge-base}};
  \node (_coref-chain-map) at (41bp,266bp) [draw,draw=none] {\textsf{:coref-chain-map}};
  \node (_triples) at (17bp,162bp) [draw,draw=none] {\textsf{:triples}};
  \node (_sentences) at (217bp,266bp) [draw,draw=none] {\textsf{:sentences}};
  \node (_knowledge-base) at (88bp,214bp) [draw,draw=none] {\textsf{:knowledge-base}};
  \node (_text) at (170bp,370bp) [draw,draw=none] {\textsf{:text}};
  \node (_ontology) at (46bp,58bp) [draw,draw=none] {\textsf{:ontology}};
  \node (_draw) at (152bp,7bp) [draw,draw=none] {\textsf{:draw}};
  \node (_reified-triples) at (88bp,110bp) [draw,draw=none] {\textsf{:reified-triples}};
  \node (_corenlp-pipeline) at (102bp,370bp) [draw,draw=none] {\textsf{:corenlp-pipeline}};
\end{scope}
\begin{scope}[>=stealth',thick,right]
  \draw [->] (_knowledge-base) ..controls (108.04bp,198.77bp) and (123.79bp,185.48bp)  .. (132bp,170bp) .. controls (148.17bp,139.51bp) and (151.55bp,98.378bp)  .. (_reified-triples-knowledge-base);
  \draw [->] (_semantic-graphs) ..controls (122.58bp,251.02bp) and (110.69bp,238.64bp)  .. (_knowledge-base);
  \draw [->] (_knowledge-base) ..controls (67.618bp,198.65bp) and (48.79bp,185.39bp)  .. (_triples);
  \draw [->] (_corenlp-annotation) ..controls (108.08bp,302.3bp) and (81.306bp,288.21bp)  .. (_coref-chain-map);
  \draw [->] (_corenlp-pipeline) ..controls (111.3bp,355.32bp) and (119.26bp,343.62bp)  .. (_corenlp-annotation);
  \draw [->] (_corenlp-annotation) ..controls (160.08bp,302.13bp) and (183.59bp,287.62bp)  .. (_sentences);
  \draw [->] (_reified-triples) ..controls (76.381bp,95.167bp) and (66.27bp,83.131bp)  .. (_ontology);
  \draw [->] (_reified-triples) ..controls (106.18bp,94.795bp) and (122.7bp,81.89bp)  .. (_reified-triples-knowledge-base);
  \draw [->] (_knowledge-base) ..controls (88bp,199.76bp) and (88bp,189.06bp)  .. (_grouped-triples);
  \draw [->] (_text) ..controls (161.64bp,356.7bp) and (153.19bp,344.28bp)  .. (_corenlp-annotation);
  \draw [->] (_grouped-triples) ..controls (88bp,147.76bp) and (88bp,137.06bp)  .. (_reified-triples);
  \draw [->] (_reified-triples-knowledge-base) ..controls (152bp,43.531bp) and (152bp,33.147bp)  .. (_draw);
  \draw [->] (_corenlp-annotation) ..controls (136bp,303.76bp) and (136bp,293.06bp)  .. (_semantic-graphs);
  \draw [->] (_coref-chain-map) ..controls (54.143bp,251.02bp) and (65.781bp,238.64bp)  .. (_knowledge-base);
\end{scope}
\end{tikzpicture}  
  \caption{Chunk annotation dependency graph}
  \label{fig:reconciler}
\end{figure}
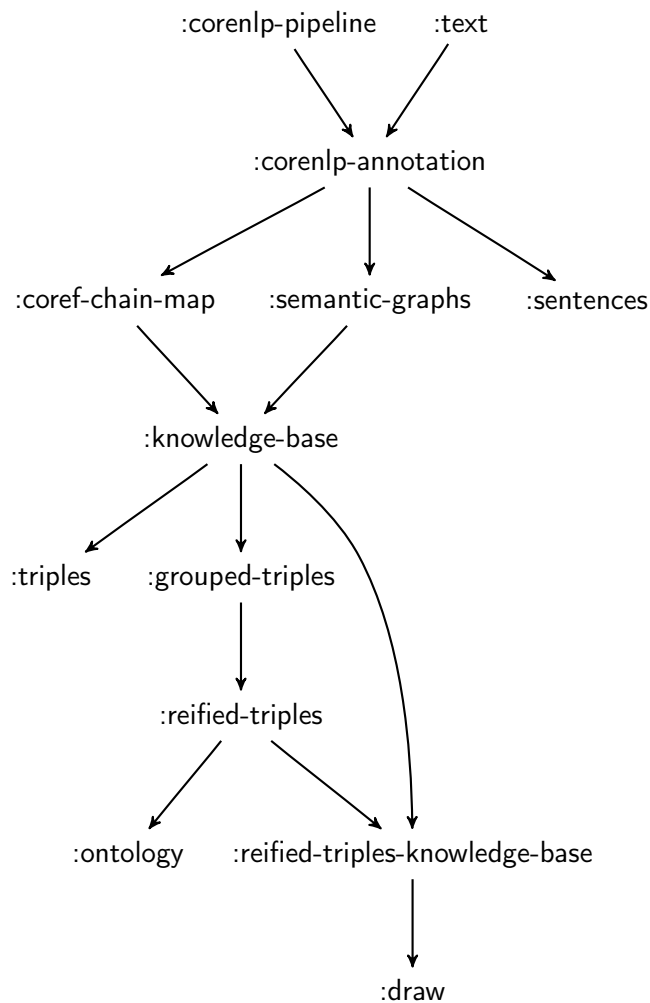

The graph in \prettyref{fig:reconciler} shows how the annotations
depend on each other. Note that graph is the central nexus of
\clidenlp that pulls all of its parts together. As such every aspect
of \clidenlp is mirrored in it.

The annotations have the following meaning:

\begin{description}
\item[:text, :corenlp-pipeline]
  The graph's inputs are a previously constructed \corenlp
  pipeline\footnote{The pipeline needs to be setup to use \corenlp's
    depedency parser and its coreference resolution system.}
  and the chunk's text.
  
\item[:corenlp-annotation]
A \corenlp annotation is created based on the input text. The
annotation provides access to all primitive NLP constructs we need
(which includes the coreference chain, the semantic graphs, all
split sentences, and information about each token).

\item[:semantic-graphs]
Extracts all semantic graphs from the annotation and creates a Clojure
representations of them. In \corenlp the class that represents nodes
in a semantic graph is called \texttt{IndexedWord}. All
\texttt{IndexedWord} instances are mapped to a word map for later use
(see \prettyref{sec:word-maps}).

\item[:coref-chain-map]
Extracts all coreference clusters from the \corenlp annotation (see
\prettyref{sec:coreference-cluster-mentions}).

\item[:sentences]
Extracts information about each of the input text's sentences from the
annotation and builds a list of sentence maps (see \prettyref{sec:sentence-maps}).

\item[:knowledge-base]
Builds a knowledge base for use with \corelogic out of the coref chain
map and semantic graphs. The process is described in \prettyref{sec:knowledge-base}.

\item[:triples]
Runs triple builders on the knowledge base that extract useful
information gathered from the semantic graphs and builds a list of
triples. The triple's subject, predicate, and object only refers to a
node in the semantic graph of one sentence. See
\prettyref{cha:triples} for more details.

\item[:grouped-triples]
Runs triple builders on the knowledge base and groups the triples'
subjects and objects by their coreference cluster. A group is a list
of coreferent or otherwise related words. In contrast with the triples
extracted by \textsf{:triples}, the grouped triple's subject and
object are a list of related words that can span multiple sentences
(the whole text) instead of only one sentence. See
\prettyref{sec:grouping-triples} for more details.

\item[:reified-triples]
Multiple grouped triples can all have the same subject or object
groups. \textsf{:reified-triples} assigns a unique name to each group
i.e. the groups are made real (reified) by giving them a name (see
\prettyref{sec:reified-triples}).

\item[:reified-triples-knowledge-base]
Adds the reified triples to the knowledge base for use by applications that
don't want to search the triples sequentially.

\item[:draw] An annotation that is introduced in detail in
  \prettyref{cha:usecase}. It tries to draw a graph specified in
  the input text and to warn about simple ambiguous sentences.
  
\item[:ontology]
  Builds an OWL ontology (see \prettyref{sec:ontology-export}).
\end{description}

An annotation is only realized when it is directly needed
or used by a dependent annotation. As a result, \clidenlp does not
always have to compute all annotations if the text changed and only does work
if it is really needed.

\begin{example}
If a client accesses the \textsf{:knowledge-base} annotation
only the following annotations are realized: \textsf{:semantic-graphs},
\textsf{:coref-chain-map}, \textsf{:corenlp-annotation},
\textsf{:text}.\footnote{Note that \textsf{:text} is the input to the
  graph and as such always realized.}
\end{example}

% \subsection{Implementation}

% \notdoing{functional, client needs to keep track of state}
% \notdoing{prismatic graphs}

% -*- TeX-master: "../master.tex"; ispell-local-dictionary: "en_US"; -*-

\section{Integration of CoreNLP}
\label{sec:corenlp}

\clidenlp tries to not rely on \corenlp's data structures, because
the data structures need to participate in \corelogic unification. Due to
the mutable nature of \corenlp's data structures extending them to
reliably support unification is problematic.

Additionally, there are inconsistencies in the usages of 0- or 1-based
indices in \corenlp's data structures. This is corrected when constructing
\clidenlp data structures and allows for easier matching up of the
different data structures based on sentence and token indices.

All data structures are implemented using Clojure records. Records are
reified maps, which compile down to Java classes. They implement
the correct interfaces, so that they can be treated as maps, which we
will do from this point on.

We introduce each record by 
\begin{itemize}
\item giving a short description of its use in \clidenlp,
\item by listing its available keys with a description of the content
  of their values,
\item and by showing examples with actual data.
\end{itemize}

% \clidenlp uses the following \corenlp components directly:
% the Stanford Parser as described in \citep{de2006,marcus1993} and the
% Stanford Deterministic Coreference Resolution System as described in
% \citep{raghunathan2010,lee2011,lee2013,recasens2013}.

\subsection{Accessing sentences of a text}
\label{sec:sentence-maps}

While accessing individual sentences of a text is not important for
the core task of \clidenlp (extracting an ontology from a text), we
need them to create \clide annotations that refer to a whole sentence
(see \prettyref{sec:clide}).

Sentence maps have the following keys:

\begin{tabular}{lp{13cm}}
  \textsf{:index} & The index of the sentence starting at 0.\\
  \textsf{:span} & The 0-based character index span $[a,b)$ of the sentence.
  The sentence starts at offset $a$ and goes up to offset $b$.\\
  \textsf{:text} &  The text of the sentence.
\end{tabular}

\begin{example}
  The text \verb|"Felix is a cat. Waldo is a dog. Tweety is a bird."|
  results in the following sentence maps:\footnote{When checking the
    spans do not forget to include the spaces between the sentences!}
  \begin{center}
\begin{multicols}{3}
\begin{tabular}{|ll|}
\hline
\textsf{:index} & 0 \\
\textsf{:span} & $[ 0 , 15 )$\\
\textsf{:text} & Felix is a cat. \\
\hline\end{tabular}
\columnbreak
\begin{tabular}{|ll|}
\hline
\textsf{:index} & 1 \\
\textsf{:span} & $[ 16 , 31 )$\\
\textsf{:text} & Waldo is a dog. \\
\hline\end{tabular}
\columnbreak
\begin{tabular}{|ll|}
\hline
\textsf{:index} & 2 \\
\textsf{:span} & $[ 32 , 49 )$\\
\textsf{:text} & Tweety is a bird. \\
\hline\end{tabular}
\columnbreak
\end{multicols}\end{center}
\end{example}

\subsection{Word maps}
\label{sec:word-maps}

Semantic graph nodes are an integral part of the triple builders
introduced in \prettyref{cha:triples}.

The semantic graph node class in \corenlp is called
\texttt{IndexedWord}. The information that an \texttt{IndexedWord}
object provides, is used to create a word map with the following keys:

\begin{tabular}{lp{13cm}}
  \textsf{:sentence} & The sentence index this word map refers to.
  This matches the sentence maps' \textsf{:index} value.\\
  \textsf{:index} & The index of the word's token starting at 1.
  \corenlp consistently starts at 1 when counting tokens.\\
  \textsf{:span} & The 0-based character index span $[a,b)$ of the
  word map. \\
  \textsf{:tag} & The word's part-of-speech tag. \\
  \textsf{:lemma} & The word's lemma.\\
  \textsf{:token} & The word's token.\\
\end{tabular}

\begin{example}
  The semantic graph for the input text
  \verb|"Felix is a cat."| has the following word maps:
\begin{center}

\begin{tikzpicture}[>=latex,line join=bevel,]
\begin{scope}[thick,inner sep=4]
  \node (node966508) at (80bp,77bp) [draw,rectangle] {\textcolor{black}{\textrm{cat} \textsf{\scriptsize NN}}};
  \node (node966509) at (23bp,7bp) [draw,rectangle] {\textcolor{black}{\textrm{Felix} \textsf{\scriptsize NNP}}};
  \node (node966511) at (127bp,7bp) [draw,rectangle] {\textcolor{black}{\textrm{a} \textsf{\scriptsize DT}}};
  \node (node966510) at (80bp,7bp) [draw,rectangle] {\textcolor{black}{\textrm{is} \textsf{\scriptsize VBZ}}};
\end{scope}
\begin{scope}[>=stealth',thick,right]
  \draw [->] (node966508) ..controls (63.065bp,66.116bp) and (53.638bp,59.558bp)  .. (47bp,52bp) .. controls (39.571bp,43.542bp) and (33.522bp,32.395bp)  .. node {\textsf{\scriptsize nsubj}} (node966509);
  \draw [->] (node966508) ..controls (92.595bp,65.698bp) and (99.863bp,58.976bp)  .. (105bp,52bp) .. controls (111.7bp,42.906bp) and (117.34bp,31.467bp)  .. node {\textsf{\scriptsize det}} (node966511);
  \draw [->] (node966508) ..controls (80bp,60.167bp) and (80bp,39.213bp)  .. node {\textsf{\scriptsize cop}} (node966510);
\end{scope}
\end{tikzpicture}

\begin{multicols}{4}
\begin{tabular}{|ll|}
\hline
\textsf{:sentence} & 0 \\
\textsf{:index} & 1 \\
\textsf{:span} & $[ 0 , 5 )$\\
\textsf{:tag} & \textsf{NNP} \\
\textsf{:lemma} & Felix \\
\textsf{:token} & Felix \\
\hline\end{tabular}
\columnbreak
\begin{tabular}{|ll|}
\hline
\textsf{:sentence} & 0 \\
\textsf{:index} & 2 \\
\textsf{:span} & $[ 6 , 8 )$\\
\textsf{:tag} & \textsf{VBZ} \\
\textsf{:lemma} & be \\
\textsf{:token} & is \\
\hline\end{tabular}
\columnbreak
\begin{tabular}{|ll|}
\hline
\textsf{:sentence} & 0 \\
\textsf{:index} & 3 \\
\textsf{:span} & $[ 9 , 10 )$\\
\textsf{:tag} & \textsf{DT} \\
\textsf{:lemma} & a \\
\textsf{:token} & a \\
\hline\end{tabular}
\columnbreak
\begin{tabular}{|ll|}
\hline
\textsf{:sentence} & 0 \\
\textsf{:index} & 4 \\
\textsf{:span} & $[ 11 , 14 )$\\
\textsf{:tag} & \textsf{NN} \\
\textsf{:lemma} & cat \\
\textsf{:token} & cat \\
\hline\end{tabular}
\columnbreak
\end{multicols}\end{center}
\end{example}

\subsection{Coreferences}
\label{sec:coreference-cluster-mentions}

\clidenlp uses \corenlp's coreference resolution system to identify
which entities in a text are similar to other entities in a text.
Coreferences are grouped in clusters. A cluster is made up of
mentions and we map them to mention maps with the following keys:

\begin{tabular}{lp{13cm}}
  \textsf{:cluster-id} & The coreference cluster id the mention map is a part of.\\
  \textsf{:sentence} & The index of the sentence that contains this
  mention map. The indices in \corenlp start at 1 here. We correct
  them to be 0-based, and as a consequence match a sentence map's
  \textsf{:index} and a word map's \textsf{:sentence} value.\\
  \textsf{:index-span} & The index span $[a,b)$ refers to the tokens starting at the
  word map with index $a$ and ends before the word map with index $b$.
  The indices are 1-based again, but we do not need to adjust them
  here, because the token indices of word maps are 1-based, too.\\
  \textsf{:text} & A clear text representation of the mention map.\\
\end{tabular}

\corenlp's coreference system provides additional information about
each mention. Information about a mention's gender, its animacy or its
number (plural or singular), is however currently unused by \clidenlp.

\begin{example} The text \verb|"Felix is a cat."| has the following
  coreference cluster and associated mention maps:
\begin{multicols}{3}
\begin{center}
\begin{tikzpicture}[>=latex,line join=bevel,scale=0.75]
\begin{scope}[thick,inner sep=4]
  \node (node966528) at (29bp,63bp) [draw,rectangle] {Felix [0:1-2]};
  \node (node966529) at (29bp,9bp) [draw,rectangle] {a cat [0:3-5]};
\end{scope}
\begin{scope}[>=stealth',thick,right]
  \draw [] (node966528) ..controls (29bp,44.496bp) and (29bp,27.551bp)  .. (node966529);
\end{scope}
\end{tikzpicture}

\end{center}\begin{tabular}{|ll|}
\hline
\textsf{:cluster-id} & 1 \\
\textsf{:sentence} & 0 \\
\textsf{:index-span} & $[ 1 , 2 )$\\
\textsf{:text} & Felix \\
\hline\end{tabular}
\begin{tabular}{|ll|}
\hline
\textsf{:cluster-id} & 1 \\
\textsf{:sentence} & 0 \\
\textsf{:index-span} & $[ 3 , 5 )$\\
\textsf{:text} & a cat \\
\hline\end{tabular}
\end{multicols}

  We see that the coreference resolution system in \corenlp identified
  \verb|Felix| to have the same meaning as \verb|a cat|.
\end{example}

\section{Building an NLP knowledge base}
\label{sec:knowledge-base}

This section describes how the data structures introduced in
\prettyref{sec:corenlp} are inserted into a \corelogic database.

\prettyref{cha:triples} makes extensive use of the knowledge base and
provides usage examples.

We first need to define the relations that we want to provide. They
are described in further detail later. \clidenlp's knowledge base
provides the following relations:

\begin{tabular}{lp{115mm}}
  \lstinline|(Xword-map $w$)| & provides access to word maps (semantic
  graph nodes).\\
  \lstinline|(depends $dep$ $reln$ $gov$)| & provides
  access to semantic graph edges.\\
  \lstinline|(same-as $w_1$ $w_2$)| & succeeds iff the word maps $w_1$
  and $w_2$ can be treated as referring to the same word.\\
\end{tabular}

Next we need to insert facts into the knowledge base. Given an input text,
\begin{enumerate}
\item for every word map $w$ of a semantic graph of a sentence
  of the text, insert the fact \lstinline|(Xword-map $w$)|.
\item for every semantic graph edge $e = (dep, reln, gov)$ of a
  semantic graph of a sentence of the text,  insert the fact
  \lstinline|(depends $dep$ $reln$ $gov$)|.
\item for the word maps $w_1$ and $w_2$ and using
  \lstinline|word-map|, \lstinline|depends| and coreference cluster
  mentions, determine if $w_1$ and $w_2$ can be treated as referring
  to the same word, then insert the facts
  \lstinline|(same-as $w_1$ $w_2$)| and
  \lstinline|(same-as $w_2$ $w_1$)|.
\end{enumerate}

\relation{(Xword-map$\ w$)}
  
\textsf{word-map} provides access to the word map $w$. It is a unary
relation. To unify with information from a word map, it needs to be used
in conjunction with \corelogic's feature extraction goal
\textsf{featurec} (see \prettyref{sec:corelogic}).

\begin{example}
  Using \corelogic's \textsf{featurec} goal, we can limit a query to only succeed with
  maps with specific features. The following query returns all
  word maps which have a tag \textsf{NN} and an index 0.
\begin{lstlisting}[]
(run* [q]
  (word-map q)
  (featurec q {:tag "NN" :index 0}))
\end{lstlisting}
\end{example}

% Every word map from a semantic graph of a sentence of an input text is
% inserted into the knowledge base.
% Word maps can be distinguished by their \textsf{:sentence} and \textsf{:index} keys.

Because matching a specific set of part of speech tags is used heavily
by the triple builders introduced in \prettyref{cha:triples},
the following helper goals are defined:

\begin{tabular}{lp{12cm}}
  \lstinline|(tago $w$ $tags$)| & suceeds iff $w$ has one of the
  tags in the vector $tags$.\\
  \lstinline|(verbo $w$)| & succeeds iff $w$ is a verb, i.e.
  if it has one of the tags \textsf{VB}, \textsf{VBD}, \textsf{VBG},
  \textsf{VBN}, \textsf{VBP} or \textsf{VBZ}.\\
  \lstinline|(nouno $w$)| & succeeds iff $w$ is a noun or
  pronoun, i.e. if it has one of the tags \textsf{NNP}, \textsf{NN},
  \textsf{NNS}. \textsf{PRP} or \textsf{PRP\$}.\\
  \lstinline|(wh-wordo $w$)| & succeeds iff $w$ is a
  \emph{wh}-word\footnote{words like e.g. \emph{what} or \emph{which}} i.e. if it has one of the tags \textsf{WDT},
  \textsf{WP}, \textsf{WP\$} or \textsf{WRB}.
\end{tabular}

\lstinline|tago| is the basis of the implementations of all of these
goals. \lstinline|tago| can be defined in the following way:
\begin{lstlisting}
(defn tago
  [$w$ $tags$]
  (fresh [$tag$]
    (word-map $w$)
    (featurec $w$ {:tag $tag$})
    (membero $tag$ $tags$)))
\end{lstlisting}

\relation{(depends$\ dependent\ relation\ governor$)}

While \textsf{word-map} provides access to semantic graph nodes,
\textsf{depends} provides access to semantic graph edges.

An edge goes from the word map $governor$ to the word map $dependent$
with the grammatical relation $relation$. We insert every edge of
every semantic graph of every sentence of a text into the knowledge
base.

$relation$ can be either a string containing a typed
dependency relation (see \citep{dependencies-manual}) or if the
relation is a collapsed relation, a vector of the first and second
part of the relation (e.g. if the relation in the semantic graph was
\textsf{prep\_of}, it gets split into the vector \lstinline|["prep", "of"]|).
Splitting a collapsed dependency in that way keeps \corelogic queries
simple by making use of its native support for unifying vectors. This
allows searching for specific prepositions or all prepositions in a
simple manner.

\begin{example}
The following query returns the governor and dependent word maps of
all edges with a prepositional relation (e.g. \textsf{prep\_of} or
\textsf{prep\_in}) from every semantic graph of a text.
\begin{lstlisting}
(run* [q]
  (fresh [dep gov p]
    (depends dep ["prep" p] gov)
    (== q [dep gov])))
\end{lstlisting}
\end{example}

\relation{(same-as$\ w_1\ w_2$)}

\lstinline!Xsame-as! asserts that the word map $w_1$ is the same as
the word map $w_2$ and that we can treat the words as being instances
of the same word group.

\lstinline!Xsame-as! should be commutative, so
that the order of $w_1$ or $w_2$ does not matter.
If \lstinline!(same-as $w_1$ $w_2$)! succeeds, \lstinline!(same-as $w_2$ $w_1$)!
succeeds, too.

Because $w_1$ and $w_2$ are instances of the same word group and we
like the word groups to be about a concrete \emph{thing}, we want
to limited them to only include pronouns, nouns, or determiners.
Including adjectives e.g. does not make sense because they are properties
of word groups. Verbs are used between two or more word groups.

\lstinline!Xsame-as! is important for grouping triples (see
\prettyref{sec:grouping-triples}). There are several aspects for when
we can consider two word maps the same.

To be included in the \lstinline!Xsame-as! relation, the word maps $w_1$
and $w_2$ need to fulfill at least one of the following rules:
\begin{itemize}
\item $w_1$ and $w_2$ need to map to the same coreference cluster.
  We need to find the corresponding word maps of each of the cluster's mentions.
  We can do this in a \corelogic query by constraining a word maps
  index to be inside of the mention's index span:
\begin{lstlisting}
(let [[start end] (:index-span mention), sentence (:sentence mention)]
  (run* [q]
    (fresh [index tag]
      (word-map q)
      (featurec q {:index index, :tag tag, :sentence sentence})
      (in index (interval start (dec end))))))
\end{lstlisting}

We further limit the query result to only include word maps that
represent pronouns, nouns, or determiners.

We run the query for every cluster mention and select every 2 combination of
the found word maps and record the facts:\footnote{While facts might be
  recorded twice, we can safely ignore this.}
\begin{lstlisting}[style=inline]
(same-as $w_1$ $w_2$)$\ \ \ \ \ \ \ \ \ \ \ \ \ \ \ \ $(same-as $w_2$ $w_1$)
(same-as $w_1$ $w_1$)$\ \ \ \ \ \ \ \ \ \ \ \ \ \ \ \ $(same-as $w_2$ $w_2$)
\end{lstlisting}

We repeat this process for every coreference cluster.

\item The query 
\begin{lstlisting}
(run* [$w_1$ $w_2$]
  (nouno $w_1$)
  (nouno $w_2$)
  (depends $w_1$ "nn" $w_2$))
\end{lstlisting}
succeeds for $w_1$ and $w_2$.

  \textsf{nn} is the \emph{noun compound modifier}
  dependency relation that asserts that one noun modifies another noun
  \citep{dependencies-manual}.

  By including compound nouns in \lstinline!Xsame-as!, we ensure that
  every word of a compound noun is assigned to the same word group
  later (see \prettyref{sec:grouping-triples}).

\item $w_1$ and $w_2$ need to be linked by a
  \emph{wh}-word. Word maps
  that are linked by a \emph{wh}-word can be found with the following
  query:
\begin{lstlisting}
(run* [$w_1$ $w_2$]
  (fresh [w]
    (wh-wordo $w_1$)
    (depends $w_1$ "nsubj" w)
    (depends w (lvar) $w_2$)))
\end{lstlisting}

\corenlp's coreference system does not include \emph{wh}-words in
its mentions. Some triples found by the triple builders in
\prettyref{cha:triples} have a \emph{wh}-word as their subject and
we need to make sure that they can be grouped together with the other
groups and do not create a group by themselves.
\end{itemize}

% -*- TeX-master: "../master.tex"; ispell-local-dictionary: "en_US"; -*-

\section{\clide annotations}
\label{sec:clide}

% \prettyref{sec:architecture} described how
% \notdoing{Annotation is used a lot all over the document: chunk
%   annotations,  NLP annotations, annotation streams, annotation types,
%   \clidenlp annotations. Try to summarize and make the differences
%   clearer? Or try to find some different words for them?}

\clide annotations are used to provide rich information about specific
parts of a text. They are currently static and
non-interactive\footnote{This means that you cannot interact with the
  annotation itself, because they are view-only. You can however
  request a new annotation.}. There is
e.g. no way to jump to a specific word in the document from an
annotation. While this limits their usefulness, it does not
prevent them from being helpful.

They follow the same model as the operations sent by \clide
(see \prettyref{sec:reconciler-text-changes}) and are
represented as a list of annotation operations, called an annotation
stream. In \clide they are simply called annotations, but we
use the term \emph{annotation stream} to distinguish them from the
annotation lists they contain.

\subsection{Annotation streams}

An annotation stream is a list of annotation operations. There are two
types of operations:
\begin{itemize}
\item \texttt{Plain($n$)}
\item \texttt{Annotate($n$, $annotations$)}
\end{itemize}
where $n$ is the annotation's length and $annotations$ is a list of
tuples $(type, content)$ with $type$ being the annotation type and
$content$ a string containing the actual annotation content.

We again translate the Scala syntax into Clojure data and replace

\begin{tabular}{rcl}
    \texttt{Plain($n$)} & with & \textsf{[:plain $n$]}\\
    \texttt{Annotate($n$, $annotations$)} & with & \textsf{[:annotate $n$ $annotations$]}\\
\end{tabular}

To apply an annotation stream we need to maintain a cursor position
starting at 0. The annotation is applied by applying each action
sequentially and mutating the cursor position afterwards.
\begin{itemize}
\item \textsf{[:plain $n$]} skips $n$ characters from the current
  cursor position $c_i$, and adds $n$ to it:\\$c_{i+1} = c_i + n$.
\item \textsf{[:annotate $n$ $annotations$]} moves the cursor $n$
  characters ahead and annotates the text span from $[c_i, c_{i+1})$
  where $c_{i+1} = c_i + n$.
\clide then interprets the annotation list $annotations$ and displays them.
\end{itemize}
where

\begin{tabular}{rl}
  $i$      & the index of the operation in the list\\
  $c_0$  & 0\\
  $c_i$ & the cursor position before applying the operation\\
  $c_{i+1}$    & the cursor position after applying the operation\\
\end{tabular}

There is a direct correspondence between annotation and
edit operations (see \prettyref{sec:reconciler}). Ignoring
$annotations$, we can treat \textsf{:plain} and
\textsf{:annotate} as \texttt{:retain} operations \citep{ring2014}.

An annotation stream should span the whole text, that is by
summing up the $n$-s of each operation, we would get the text length.

\clide defines several annotation types. \clidenlp uses the following
subset of them:
\begin{description}
\item[\textsf{Class}] sets the CSS class to use for the annotation.
\item[\textsf{Tooltip}] sets the tooltip used when hovering over the
  annotation.
\item[\textsf{WarningMessage}] display a warning inline.
\item[\textsf{Output}] for displaying generated information, that is
  not displayed inline by default.
\end{description}

\clide allows HTML inside of its annotations, which means that we can
display richer annotations than only simple plain text annotations.

%\vspace*{1em}
\begin{example}
  Given the text \lstinline|"The cat is hungry."| of length 18 and the annotation stream

\textsf{[[:plain $4$], [:annotate $3$ [[:Class \lstinline|"error"|]]],
  [:plain $11$]]}

and assuming \textsf{[:Class \lstinline|"error"|]} is meant to color
the annotated text red, applying the annotation stream to the text results in

\lstinline|"The $\textsf{\textcolor{red}{cat}\textvisiblespace}$is$\textsf{\textvisiblespace}$hungry."|
\end{example}

\subsection{Annotation levels}
\label{sec:clidenlp-annotation-levels}

While \clide annotations can annotate arbitrary ranges of a text, it
is useful to distinguish between different annotation levels in \clidenlp:
\begin{description}
\item[Text] Annotates the whole text
\item[Chunk] Annotates a chunk (as defined in \prettyref{sec:reconciler})
\item[Word] Annotates a single word
\item[Sentence] Annotates a whole sentence
\end{description}

Every annotation in \clidenlp, with the exception of a text level
annotation, is relative to a chunk. By doing so, we keep the
annotation creation as simple as possible.

As discussed in \prettyref{sec:reconciler}, a chunk has an associated
span that indicates the chunk's text position inside of the global
text. A chunk's internal span begins at 0. \corenlp never sees the
whole text at once, but instead only sees the texts of every chunk
separately, so all spans returned by \corenlp also begin at
0.\footnote{After some correction (see \prettyref{sec:corenlp})}
It follows that the NLP knowledge base is chunk local, too.
As the annotations created by \clidenlp all use \corenlp annotations
or the NLP knowledge base, we ideally should use chunk local offsets when
creating annotation streams. We later project their chunk local offsets to
offsets that \clide can interpret correctly.

\subsection{Annotations provided by \clidenlp}
\label{sec:clidenlp-annotations}

In this section we describe each annotation that \clidenlp provides by
showing an example of how they appear in \clide. The annotations make
use of one more of the chunk annotations as described in
\prettyref{sec:reconciler}. There might be some overlap in their
names, but they should be treated as separate entities. The names are
presented to the user by \clide, who can enable them individually (see
the names to the left with the ``eyes'' in \prettyref{fig:screen-semantic-graph}).

%% for the images here, use a width of `image_width / 2`px to not get
%% fuzzy images
\annotation{chunk-separators}{Text}{
  Highlights the chunk separators. Chunks are described in
  \prettyref{sec:reconciler}. This makes the chunkers decisions visible. 
}{\includegraphics[width=117px]{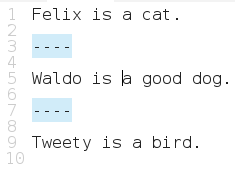}}

\annotation{semantic-graph}{Sentence}{
  Highlights a sentence and displays the sentence's semantic graph.
  This annotation is highly useful when creating a new triple builder.
}{\includegraphics[width=224px]{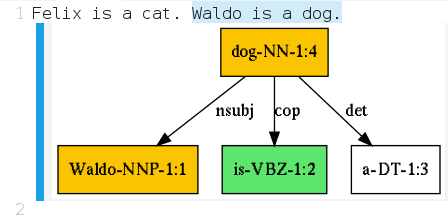}}

\annotation{coref-cluster}{Chunk}{
  Shows the coref clusters that \corenlp found for the selected chunk's text.
}{\includegraphics[width=169px]{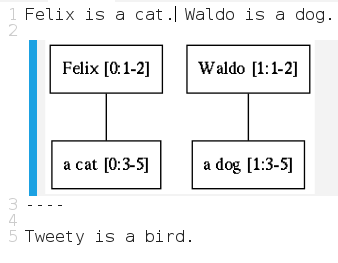}}

\annotation{same-as}{Word}{
  A direct interface to the knowledge base that uses the
  \textsf{same-as} relation to highlight the related words of the
  selected word.
}{\includegraphics[width=168px]{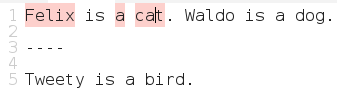}}

\annotation{triples}{Chunk}{
  Shows the raw triples extracted from running the triple builders on
  the selected chunk's text.
}{\includegraphics[width=206px]{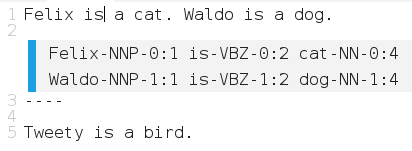}}

\annotation{grouped-triples}{Chunk}{
  Shows the triples' groups.
}{\includegraphics[width=438px]{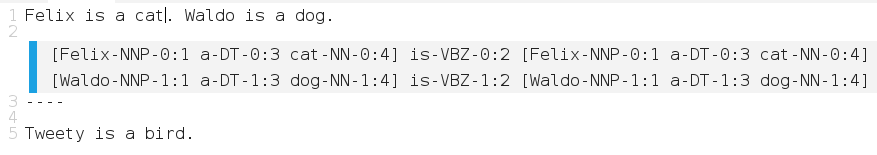}}

\annotation{reified-triples}{Chunk}{
  Shows a table with all reified triples extracted from the selected
  chunk's text.
}{\includegraphics[width=196px]{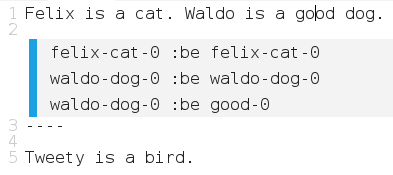}}

\annotation{reified-name}{Word}{
  This annotation is identical to the \textsf{same-as} annotation, but
  additionally adds a tooltip to the highlighted words, showing the word group
  they belong to.
}{\includegraphics[width=198px]{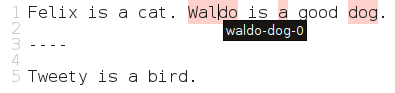}}

% -*- TeX-master: "../master.tex"; ispell-local-dictionary: "en_US"; -*-

\chapter{Triples}
\label{cha:triples}

In the previous chapter we built an NLP knowledge base.
This chapter introduces triple builders that runs simple \corelogic
queries on it and extracts triples. %the knowledge base we built in the previous chapter.
Triples have the form $(\textit{subject}, \textit{predicate}, \textit{object})$.
They represent a unit of useful information.

Where possible a triple's subject and object each have \emph{one}
corresponding word map in a semantic graph. Using the knowledge base
introduced in the previous chapter, we group the triple's subject and
object with other related subjects or objects of other triples,
yielding an ontology.

\section{Triple builders}
\label{sec:triple-builders}

Triple builders extract triples. Triples have the form
$(\textit{subject}, \textit{predicate}, \textit{object})$.
They represent a unit of useful information that is directly extracted
from the semantic graph.

A subject usually does something (predicate) with an
object. Every subject and object is directly linked to a
word map. For predicates however this is not always possible, because
there are predicates which are implied by the semantic graph (see the triple builders
\textsf{nsubj-amod} or \textsf{possessive} for an example of this).
These predicates are called \emph{derived} and written with a colon
prefix (e.g. \textsf{:be}) to distinguish them from word map predicates.

While the triple builders only extracts information from semantic
graphs, they could be extended to include information from ontologies
like WordNet \citep{wordnet}, VerbOcean \citep{verbocean} or DBpedia
\citep{dbpedia} to further constrain the triples that are found.

A triple builder is implemented as a \corelogic goal.\footnote{Note
  that we deviate from the \corelogic convention of marking a goal
  with a postfix $^o$ here.} It either
succeeds with a triple or it does not. All triple builders are
tied together in a logical disjunction (\lstinline|conde|) and are all
tried in turn. Most of the triple builder's names are directly derived
from the semantic graph relation's names they use.

We show each triple builder with
\begin{itemize}
\item the actual \corelogic query,
\item an example input sentence,
\item the corresponding semantic graph for the input sentence, with
  highlights for the relevant nodes and edges to make it easier to
  follow along,
\item a table of the triples that it found when run on the input
  sentence, displaying the lemmas of its subject, predicate or object,
\item and a discussion of the triple builder.
\end{itemize}
Note that a triple builder may succeed multiple times and thus may
find more than one triple.

\begin{triplebuilder}{nsubj-amod}{The hungry cat eats.}
  {:edge-exclusions #{'nsubj' ['conj' 'and']}}
The meaning of a subject can be modified with an adjectival modifier
(\textsf{amod}) \citep{dependencies-manual}.

This triple builder captures the fact that \emph{cat} refers not just
to cat but to a \emph{hungry} cat. Because there is no node in the
graph that can take the role of a predicate, we introduce a derived
predicate \textsf{:be} instead.
\end{triplebuilder}

\begin{triplebuilder}{nsubj-pred-dobj}{It eats the mouse.}{}
  This triple builder captures facts from one of the simplest kind of
  sentences. A subject directly connected to an object with a
  predicate.
\end{triplebuilder}

\begin{triplebuilder}{nsubj-VB}{The cat eats.}{}
  Some sentences have intransitive verbs\footnote{verbs with no
    object} and no object. However, they might still contain useful
  information.

  Here we capture the fact that the \emph{cat} \emph{eats}.
  To do this, the verb becomes our triple object and we use a derived
  predicate \textsf{:be}. This mirrors the behavior used in e.g. the
  triple builder \textsf{nsubj-amod} for adjectives.

  We limit ourselves to sentence with verbs that are
  in the past tense (tag \textsf{VBD}) or singular present (tags
  \textsf{VBP} and \textsf{VBZ}) to not always trigger this
  triple builder for all sentences with an \textsf{nsubj} relation
  between a noun and a verb.
\end{triplebuilder}

\begin{triplebuilder}{nsubj-adj-cop}{The cat is often full.}{}
  This triple builder captures adjectives of subjects, e.g. that the
  \emph{cat is full}.

  We can qualify the object \emph{full} with an additional triple. See
  the next triple builder \textsf{nsubj-advmod}.
\end{triplebuilder}

\begin{triplebuilder}{nsubj-advmod}{The cat is often full.}{}
  In combinations with \textsf{nsubj-adj-cop} finds additional
  descriptions of an adjective, but also additional properties of
  subjects.

  Following e.g. the triple builders \textsf{nsubj-amod} or
  \textsf{nsubj-VB}, \textsf{:be} is used as our predicate again.

  The object from the triple found in the example of
  \textsf{nsubj-adj-cop} \emph{full} is qualified here with
  \emph{often}.
\end{triplebuilder}

\begin{triplebuilder}{nsubj-pred-acomp}{It looks hungry.}{}
  Verbs and adjectives have an \textsf{acomp} (adjectival complement)
  relation, if an adjective can be treated as the verb's object
  \citep{dependencies-manual}.

  The triple extraction is straightforward. The adjective becomes the
  triple's object. The verb's subject the triple's subject and the
  verb itself the predicate.
\end{triplebuilder}

\begin{triplebuilder}{nsubj-pred-xcomp}{He managed to enter the
    house.}{}
  \emph{enter} is an open clausal complement (\textsf{xcomp}) of
  \emph{managed}. \emph{enter} does not have its own subject but
  refers to the subject of \emph{managed} (\emph{He})
  \citep{dependencies-manual}.

  There are two useful triples that can be extracted from the graph. One is
  the fact that \emph{He} \emph{managed} to do something
  (\emph{enter}) and one is a fact about what \emph{He} entered (the
  \emph{house}).
\end{triplebuilder}

\begin{triplebuilder}{nsubjpass-pred-agent}{He is swayed by a
    warning.}{}
  An \textsf{agent} is ``introduced by the preposition \emph{by}''
  \citep{dependencies-manual}. This also implies a passive subject (\emph{he}),
  which we will use for our triple's object. The agent (\emph{warning})
  is used as the triple's subject, because it does something with the
  passive subject.
\end{triplebuilder}

\begin{triplebuilder}{agent-ccomp-dobj}
  {It is known by X that you do Y}
  {:reverse-nodes?  true,
   :nodesep         '0.13'
   :edge-inclusions #{'ccomp'}}
This triple builder extracts the fact that \emph{X} talks about
\emph{Y}. Because there is no direct graph node we could use for the
predicate here, we introduce another derived predicate
\textsf{:about}.

Any other information from this sentence, will be captured by other
triple builders.
\end{triplebuilder}

\begin{triplebuilder}{possessive}{John's house is
    red.}{:edge-exclusions #{'nsubj'}}
  We want to determine what kind of possessions a subject has.
  The semantic graph has the relation \emph{poss} (possession
  modifier) for this. Because we are missing a direct predicate in the
  graph, we introduce a derived predicate \textsf{:have}.
\end{triplebuilder}

\begin{triplebuilder}{nsubjpass-ccomp}
   {X and Y are connected as are B and C}
  {:reverse-nodes?  true
   :nodesep         '0.08'}
 The query searches for nouns that are connected via some predicate
 and have a clausal complement (\emph{ccomp}) or passive nominal subject
 (\emph{nsubjpass}) relation with it.

 Because the subject/object-order of the nouns in the resulting triple
 shouldn't matter, the query is allowed to succeed for all possible
 combinations of them,  with the additional constraint that the nouns
 have to be connected via some conjunction. This prevents extracting
 wrong triples, like e.g. ``b connect x''.

 \begin{counterexample}
   This triple builder sometimes finds information that is obviously
   wrong. If we vary the sentence a little bit by changing
   ``connected'' to ``proven'',  we get essentially the same semantic
   graph and triples with wrong information.

   Running the triple builder on the sentence ,,X and Y are proven as
   are B and C'' would return the triples

   {\footnotesize
     \begin{tabular}{l|l|l}y & prove & x \\x & prove & y \\c & prove & b \\b & prove & c \\\end{tabular}}\\\ \\
   Clearly this is wrong, ``proven'' is an intransitive
   verb here and X and Y did not ``prove'' each other, but were proven by some
   (in this case unknown) agent.

   % \notdoing{add citation for definition of transitive/intransitive}
   % The issue gets worse if we consider that verbs can be either
   % intransitive or transitive depending on their context.

   There is something missing here. Integrating a verb
   ontology might help to determine if a verb is transitive or intransitive.
   % One potential solution to this problem is to integrate a verb
   % ontology into \clidenlp to check if the verb is ambitransitive,
   % transitive, or intransitive and handling every case appropriately.

   In the case of intransitive verbs a more appropriate result
   (matching the triples found by the triple builder \textsf{nsubj-VB}) would be:

   {\footnotesize
   \begin{tabular}{l|l|l}
     x & \textsf{:be} & prove\\
     y & \textsf{:be} & prove\\
     c & \textsf{:be} & prove\\
     b & \textsf{:be} & prove\\
   \end{tabular}}
 \end{counterexample}
\end{triplebuilder}

\begin{triplebuilder}{prep-noun}{They have a distance of 2 cm.}{:edge-exclusions #{'dobj'}}
  This triple builder captures preposition between two nouns (or
  pronouns). Because CoreNLP collapses prepositions, we introduce
  derived predicates for each preposition. In this case because of the
  edge between the subject and object (\textsf{prep\textunderscore
    of}) the predicate will be \textsf{:of}.
\end{triplebuilder}

\begin{triplebuilder}{noun-prep-noun}{X is in the same state as Y}{}
The complement to \textsf{prep-noun} that captures preposition that
are indirectly connected to a noun.

Like in \textsf{prep-noun} we again use a derived predicate for
capturing the preposition. Because the nouns are not connected
directly, we let the query succeed twice. Once for capturing the
preposition as our predicate (\textsf{:in}) and once for capturing
the actual predicate from the graph (\emph{is}).
\end{triplebuilder}

\begin{triplebuilder}{noun-num}{He has 5 apples.}{:edge-exclusions #{'dobj'}}
  This triple builder captures numeric modifiers (\textsf{num})
  \citep{dependencies-manual} of nouns, e.g. how many of \emph{apple}s
  there are (\emph{5}).

  Because we are missing a direct predicate in the graph, we will use
  the derived predicate \textsf{:be}.
\end{triplebuilder}

\begin{triplebuilder}{advmod-npadvmod-num}{A is 10 cm wide.}{}
The complement to \textsf{noun-num} to capture the unit of a
measurement.
\end{triplebuilder}

\prettyref{tab:triple-builders} shows all triples that are captured by
running them on an example text that is used in
\prettyref{cha:usecase} to draw a graph from the text.

Looking at the table it is clear that all necessary information for
this is available, however we currently do not have any way of
discerning if an instance of e.g. ``distance'' is the same as another
instance of ``distance'' or what ``it'' refers to.

\section{Reifying triples}
\label{sec:grouping-triples}
\label{sec:reified-triples}
Currently the triples' subjects or objects refer to one word map only.
For example, looking at \prettyref{tab:triple-builders} we see several
subjects with a ``Test'' or ``it'' lemma. There is no way that we can
know if the instances refer to the same entity or to distinct
entities.

By using the \textsf{same-as} relation we build in
\prettyref{sec:knowledge-base}, we can group the triples' subjects and
objects with all other words in the knowledge base that can be treated
as refering to the same entity, reifying the triples.

%% We can see that ``it'' and ``test'' belong to the cluster with ``Edge
%% Test''. 

If we apply \textsf{same-as} to every subject and object of our
triples our table might look like \prettyref{tab:grouped-triples},
where the subjects and objects are replaced by the word groups that
the original word belonged to.

\begin{example} Let us look at the coreference clusters found by
  \corenlp for our input text. This gives us an approximation of what
  the \textsf{same-as} relation looks like:
\begin{center}
\begin{tikzpicture}[>=latex,line join=bevel,scale=0.75]
\begin{scope}[thick,inner sep=4]
  \node (node607622) at (456bp,171bp) [draw,rectangle] {Node Noname [2:1-3]};
  \node (node607621) at (313bp,117bp) [draw,rectangle] {node End [3:4-6]};
  \node (node607618) at (177bp,63bp) [draw,rectangle] {Node Start [3:1-3]};
  \node (node607619) at (177bp,9bp) [draw,rectangle] {node Start [3:10-12]};
  \node (node607620) at (313bp,171bp) [draw,rectangle] {node End [0:5-7]};
  \node (node607614) at (40bp,117bp) [draw,rectangle] {Edge Test [4:1-3]};
  \node (node607615) at (40bp,63bp) [draw,rectangle] {Edge Test [0:1-3]};
  \node (node607616) at (177bp,171bp) [draw,rectangle] {node Start [2:4-6]};
  \node (node607617) at (177bp,117bp) [draw,rectangle] {node Start [0:10-12]};
  \node (node607623) at (456bp,117bp) [draw,rectangle] {node Noname [3:13-15]};
  \node (node607613) at (40bp,171bp) [draw,rectangle] {It [1:1-2]};
\end{scope}
\begin{scope}[>=stealth',thick,right]
  \draw [] (node607622) ..controls (456bp,152.5bp) and (456bp,135.55bp)  .. (node607623);
  \draw [] (node607616) ..controls (177bp,152.5bp) and (177bp,135.55bp)  .. (node607617);
  \draw [] (node607618) ..controls (177bp,44.496bp) and (177bp,27.551bp)  .. (node607619);
  \draw [] (node607620) ..controls (313bp,152.5bp) and (313bp,135.55bp)  .. (node607621);
  \draw [] (node607614) ..controls (40bp,98.496bp) and (40bp,81.551bp)  .. (node607615);
  \draw [] (node607613) ..controls (40bp,152.5bp) and (40bp,135.55bp)  .. (node607614);
  \draw [] (node607617) ..controls (177bp,98.496bp) and (177bp,81.551bp)  .. (node607618);
\end{scope}
\end{tikzpicture}
\end{center}

If we look at the subject word map $w$ of the triple

  \begin{tabular}{lll}
    Test & go & End
  \end{tabular}

  And we search for every $v$ for which \lstinline|(same-as $w$ $v$)|
  holds, we get the word group:\footnote{The numbers correspond to the
    word maps sentence and token index ($sentence$, $index$).}

  Edge (0,1), Test (0,2), it (1,1), Edge (4,1)
\end{example}

Because working with a list of word maps is cumbersome
and also hard to refer to, we give each word group a unique name. A word
group's name is made up of the nouns (this excludes \emph{Wh}-words or
pronouns) of each word in the group. Because some names may not be
unique, we add a number to it. We increment the number every time there is a
duplicate name for a group.

A predicate can be derived or refer to a
concrete word map. We simply reify the predicates by making them all
derived. For non-derived predicates we use it's lemma and derived predicates
are copied verbatim. This essentially also groups the predicates
together.

The result of applying these steps to \prettyref{tab:grouped-triples}
can be seen in \prettyref{tab:reified-triples}.

\begin{example} There are three word groups with the same name:
  \textsf{cm-0}, \textsf{cm-1}, \textsf{cm-2}

  Because they all refer to separate entities, they all have a unique
  number.
\end{example}

\begin{example} In \prettyref{tab:grouped-triples} the predicate
  \emph{connect} occurs several times as (potentially distinct) word maps. In
  \prettyref{tab:reified-triples} we reduced them all to a single
  \textsf{:connect}.
\end{example}

To allow clients to access the reified triple, we update the knowledge
base from \prettyref{sec:knowledge-base} with a new relation:

\relation{(triple$\ t$)}

\textsf{triple} is a unary relation and we simply add every triple we
found to it. Clients can then access any information from the triple
by unifying with $t$. A reified triple is a map that has the following schema:
\begin{lstlisting}[style=inline]
{$\textsf{:subject}$ {$\textsf{:symbol}$ $\textrm{\emph{word group name}}$
$\ \ \ \ \ \ \ \ \ \ \ \ \,$ $\textsf{:group}$ $\textrm{\emph{vector of word maps}}$}
$\ \,$$\textsf{:predicate}$ $\textrm{\emph{predicate}}$
$\ \,$$\textsf{:object}$ $\textrm{\emph{same layout as \textsf{:subject}}}$}
\end{lstlisting}

%% Triple reification does not throw away any information.
%% As the schema above shows a triple's subject or object word group can
%% still be accessed. The original triple as found by a triple builder
%% can be accessed in the triple's metadata through the
%% \textsf{:origin} key. \notdoing{Explain metadata in basics} The origin is
%% not available directly because we do not want the origin to have any
%% influence on triple equality\footnote{Two triples might have the same
%%   subject, predicate and object but have different origins. }.

\begin{table}[p]
  \caption[Result of running all triple builders on an example text]
  {Result of running all triple builders on the input text:\\
    \texttt{Edge Test goes to node End and starts at node Start.
      It is 15 cm long.
      Node Noname and node Start have a distance of 10 cm.
      Node Start and node End are connected as are node Start and node
      Noname. Edge Test is 4 cm long.}\\
  Each row shows a captured triple with its subject, predicate and
  object and the triple builder that created it.}
\label{tab:triple-builders}
\centering
\begin{tabular}{lcll}
 \bf Subject & \bf Predicate & \bf Object & \bf Triple builder\\\hline
Start & have & distance & \textsf{:nsubj-pred-dobj}\\
Noname & have & distance & \textsf{:nsubj-pred-dobj}\\
Start & have & distance & \textsf{:nsubj-pred-dobj}\\
distance & \textsf{:of} & cm & \textsf{:prep-noun}\\
long & \textsf{:be} & cm & \textsf{:advmod-npadvmod-num}\\
cm & \textsf{:be} & 4 & \textsf{:noun-num}\\
Test & go & End & \textsf{:noun-prep-noun}\\
Test & \textsf{:to} & End & \textsf{:noun-prep-noun}\\
long & \textsf{:be} & cm & \textsf{:advmod-npadvmod-num}\\
cm & \textsf{:be} & 10 & \textsf{:noun-num}\\
Test & start & Start & \textsf{:noun-prep-noun}\\
Test & \textsf{:at} & Start & \textsf{:noun-prep-noun}\\
cm & \textsf{:be} & 15 & \textsf{:noun-num}\\
Test & be & long & \textsf{:nsubj-advmod}\\
it & be & long & \textsf{:nsubj-advmod}\\
Noname & connect & Start & \textsf{:nsubjpass-ccomp}\\
Start & connect & Noname & \textsf{:nsubjpass-ccomp}\\
End & connect & Start & \textsf{:nsubjpass-ccomp}\\
Start & connect & End & \textsf{:nsubjpass-ccomp}\\
Test & \textsf{:be} & be & \textsf{:nsubj-VB}\\
Noname & \textsf{:be} & have & \textsf{:nsubj-VB}\\
it & \textsf{:be} & be & \textsf{:nsubj-VB}\\
Test & \textsf{:be} & start & \textsf{:nsubj-VB}\\
Start & \textsf{:be} & have & \textsf{:nsubj-VB}\\
Test & \textsf{:be} & go & \textsf{:nsubj-VB}\\
\end{tabular}
\end{table}

\begin{table}[p]
  \caption{Grouped triples based on \prettyref{tab:triple-builders}}
\label{tab:grouped-triples}
\centering
\begin{tabular}{p{65mm}cp{65mm}}
 \bf Subject word group & \bf Predicate & \bf Object word group\\\hline
\small node (0,10), Start (0,11), node (2,4), Start (2,5), Node (3,1), Start (3,2), node (3,10), Start (3,11) & have & \small distance (2,8) \\
\small Node (2,1), Noname (2,2), node (3,13), Noname (3,14) & have & \small distance (2,8) \\
\small node (0,10), Start (0,11), node (2,4), Start (2,5), Node (3,1), Start (3,2), node (3,10), Start (3,11) & have & \small distance (2,8) \\
\small distance (2,8) & \textsf{:of} & \small cm (2,11) \\
\small long (4,6) & \textsf{:be} & \small cm (4,5) \\
\small cm (4,5) & \textsf{:be} & \small 4 (4,4) \\
\small Edge (0,1), Test (0,2), it (1,1), Edge (4,1), Test (4,2) & go & \small node (0,5), end (0,6), node (3,4), end (3,5) \\
\small Edge (0,1), Test (0,2), it (1,1), Edge (4,1), Test (4,2) & \textsf{:to} & \small node (0,5), end (0,6), node (3,4), end (3,5) \\
\small long (1,5) & \textsf{:be} & \small cm (1,4) \\
\small cm (2,11) & \textsf{:be} & \small 10 (2,10) \\
\small Edge (0,1), Test (0,2), it (1,1), Edge (4,1), Test (4,2) & start & \small node (0,10), Start (0,11), node (2,4), Start (2,5), Node (3,1), Start (3,2), node (3,10), Start (3,11) \\
\small Edge (0,1), Test (0,2), it (1,1), Edge (4,1), Test (4,2) & \textsf{:at} & \small node (0,10), Start (0,11), node (2,4), Start (2,5), Node (3,1), Start (3,2), node (3,10), Start (3,11) \\
\small cm (1,4) & \textsf{:be} & \small 15 (1,3) \\
\small Edge (0,1), Test (0,2), it (1,1), Edge (4,1), Test (4,2) & be & \small long (4,6) \\
\small Edge (0,1), Test (0,2), it (1,1), Edge (4,1), Test (4,2) & be & \small long (1,5) \\
\small Node (2,1), Noname (2,2), node (3,13), Noname (3,14) & connect & \small node (0,10), Start (0,11), node (2,4), Start (2,5), Node (3,1), Start (3,2), node (3,10), Start (3,11) \\
\small node (0,10), Start (0,11), node (2,4), Start (2,5), Node (3,1), Start (3,2), node (3,10), Start (3,11) & connect & \small Node (2,1), Noname (2,2), node (3,13), Noname (3,14) \\
\small node (0,5), end (0,6), node (3,4), end (3,5) & connect & \small node (0,10), Start (0,11), node (2,4), Start (2,5), Node (3,1), Start (3,2), node (3,10), Start (3,11) \\
\small node (0,10), Start (0,11), node (2,4), Start (2,5), Node (3,1), Start (3,2), node (3,10), Start (3,11) & connect & \small node (0,5), end (0,6), node (3,4), end (3,5) \\
\small Edge (0,1), Test (0,2), it (1,1), Edge (4,1), Test (4,2) & \textsf{:be} & \small be (4,3) \\
\small Node (2,1), Noname (2,2), node (3,13), Noname (3,14) & \textsf{:be} & \small have (2,6) \\
\small Edge (0,1), Test (0,2), it (1,1), Edge (4,1), Test (4,2) & \textsf{:be} & \small be (1,2) \\
\small Edge (0,1), Test (0,2), it (1,1), Edge (4,1), Test (4,2) & \textsf{:be} & \small start (0,8) \\
\small node (0,10), Start (0,11), node (2,4), Start (2,5), Node (3,1), Start (3,2), node (3,10), Start (3,11) & \textsf{:be} & \small have (2,6) \\
\small Edge (0,1), Test (0,2), it (1,1), Edge (4,1), Test (4,2) & \textsf{:be} & \small go (0,3) \\
\end{tabular}
\end{table}

\begin{table}[p]
  \caption{Reified triples based on \prettyref{tab:grouped-triples}}
\label{tab:reified-triples}
\centering
\begin{tabular}{lcl}
 \bf Subject & \bf Predicate & \bf Object \\\hline
\textsf{node-start-0} & \textsf{:have} & \textsf{distance-0} \\
\textsf{node-noname-0} & \textsf{:have} & \textsf{distance-0} \\
\textsf{distance-0} & \textsf{:of} & \textsf{cm-2} \\
\textsf{long-0} & \textsf{:be} & \textsf{cm-1} \\
\textsf{cm-1} & \textsf{:be} & \textsf{num-4-0} \\
\textsf{edge-test-0} & \textsf{:go} & \textsf{end-node-0} \\
\textsf{edge-test-0} & \textsf{:to} & \textsf{end-node-0} \\
\textsf{long-1} & \textsf{:be} & \textsf{cm-0} \\
\textsf{cm-2} & \textsf{:be} & \textsf{num-10-0} \\
\textsf{edge-test-0} & \textsf{:start} & \textsf{node-start-0} \\
\textsf{edge-test-0} & \textsf{:at} & \textsf{node-start-0} \\
\textsf{cm-0} & \textsf{:be} & \textsf{num-15-0} \\
\textsf{edge-test-0} & \textsf{:be} & \textsf{long-0} \\
\textsf{edge-test-0} & \textsf{:be} & \textsf{long-1} \\
\textsf{node-noname-0} & \textsf{:connect} & \textsf{node-start-0} \\
\textsf{node-start-0} & \textsf{:connect} & \textsf{node-noname-0} \\
\textsf{end-node-0} & \textsf{:connect} & \textsf{node-start-0} \\
\textsf{node-start-0} & \textsf{:connect} & \textsf{end-node-0} \\
\textsf{edge-test-0} & \textsf{:be} & \textsf{be-1} \\
\textsf{node-noname-0} & \textsf{:be} & \textsf{have-0} \\
\textsf{edge-test-0} & \textsf{:be} & \textsf{be-0} \\
\textsf{edge-test-0} & \textsf{:be} & \textsf{start-0} \\
\textsf{node-start-0} & \textsf{:be} & \textsf{have-0} \\
\textsf{edge-test-0} & \textsf{:be} & \textsf{go-0} \\
\end{tabular}
\end{table}

% \section{A note on providing temporal information for triples}

% \notdoing{move to conclusion?}
% \notdoing{remove completely?}
% \notdoing{if not removed, need to sketch this out some more}
% \notdoing{there are a lot more tenses than the 3 mentioned here (with
%   wrong names, too) \dots}

% Currently all triple builder ignore the tenses of verbs. It might be
% possible to add \lstinline!verbo!constraints to all triple builders,
% and then redefining the meaning of the constraint so that the triple
% builders only capture present tense (or past tense, or future tense) verbs.

% If we do this for all tenses we would get multi-layer triples,
% where each layer represents a different state of the triples.

% The text ``The house was red. It is now green. It will be yellow.'' would create three
% layers:

% past tense layer at $t_{-1}$: house-0 :be red-0

% present tense layer at $t_0$: house-0 :be green-0

% future tense layer at $t_1$: house-0 :be yellow-0

% \subsection{Updating the knowledge base}

\newpage
\section{Exporting an OWL ontology}
\label{sec:ontology-export}
\label{sec:ontology}

We create an OWL class for every subject and object word group of a
triple. OWL 2 introduces a feature called \emph{punning}
\citep{punning} where we create an individual for each of our classes
and then use object properties on these individuals to describe
relationships between the classes. 

Because a triple's predicate describes a relationship between its
subject and object, we use an object property with a name based on the
predicate and add an axiom to the ontology that links the subject's
individual with the object's individual via this object property.

A subject or object word group contains word maps that have
additional information about that group, such as all of the actual
words (tokens) that make up the group. We add this information as
datatype properties to the subject's or object's individual.

\begin{figure}[h!]
\centering
\includegraphics[width=\textwidth]{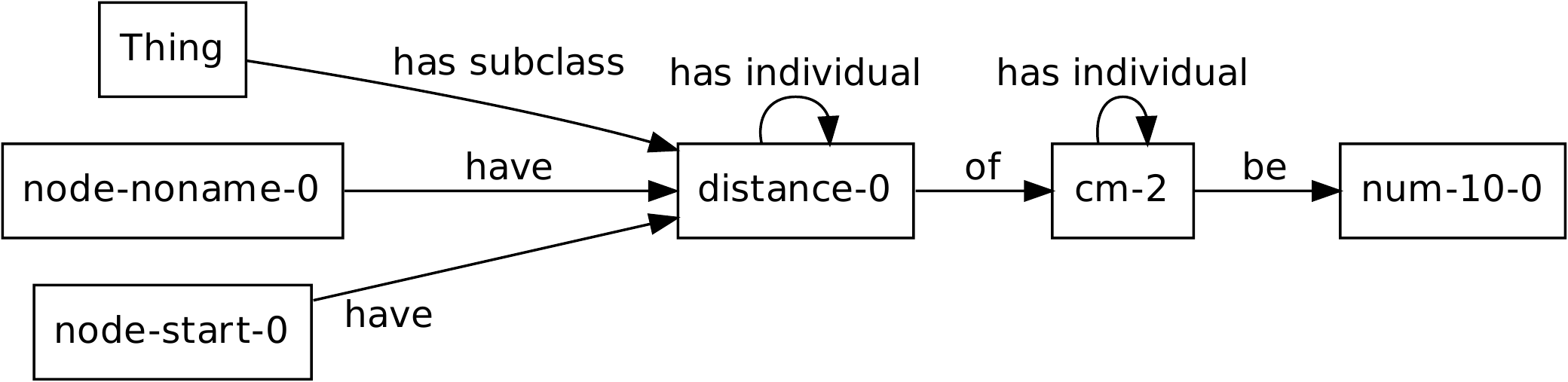}
\caption{A view on the ontology that is extracted from \prettyref{tab:reified-triples}}
\label{fig:ontology}
\end{figure}

\prettyref{fig:ontology} shows a subset of an ontology that is
based on \prettyref{tab:reified-triples}. The ``has
individual'' loops are an artifact of our use of punning. Even though
\textsf{node-start-0}, \textsf{node-noname-0} and \textsf{num-10-0}
are only shown as individuals, they are still represented as classes
(and as subclasses of \textsf{Thing}) in the full ontology.

\vspace*{-1em}
\begin{example} We can query the ontology using SparQL. E.g. to return all
  subject word groups with their constituent tokens that are linked to the word
  group \textsf{distance-0}, we can run the following query:
\begin{lstlisting}[language=sparql]
PREFIX : <$\textsf{http://clide.informatik.uni-bremen.de/clide-nlp\#}$>
SELECT ?subject ?token WHERE {
 ?subject :have $\textsf{:distance-0}$.
 ?subject $\textsf{:hasToken}$ ?token
}
\end{lstlisting}

Query result:

\begin{tabular}{ll}
?subject & ?token \\
\hline
\sf node-noname-0 & ``Noname''\\
\sf node-noname-0 & ``Node''\\
\sf node-noname-0 & ``node''\\
\sf node-start-0 & ``Start''\\
\sf node-start-0 & ``node''\\
\end{tabular}

\end{example}

% -*- TeX-master: "../master.tex"; ispell-local-dictionary: "en_US"; -*-
\chapter{Use case: Graph creation from a natural language
  specification}
\label{cha:usecase}

This chapter introduces an example application that can create
graphs from a text. The application is directly integrated into
\clidenlp and uses the reified triples introduced in
\prettyref{sec:reified-triples} as its input.

The application and its input can be seen in the reconciler's
dependency graph \prettyref{fig:reconciler} and is named
\textsf{:draw} there.

The output is a graph with the nodes and edges as
described in the text and a list of warnings about ambiguous or
incomplete information extracted from the triples.

The application should detect
\begin{itemize}
\item named nodes,
\item edges that are directly specified with a name,
\item edges that are indirectly specified as a connection
  between two nodes, and
\item edge lengths.
\end{itemize}

Emit warnings when there are
\begin{itemize}
\item simple contradictions, like different lengths for the same edge,
\item unfinished edge or node specifications,
\item sentences that specify the same node or edge twice, and
\item non-integer edge lengths.
\end{itemize}

To achieve these goals, we must specify what kind of sentences we
would like to understand. We can specify edges with the
following sentences:

\begin{center}
\begin{tabular}{l|l}
  \bf Type & \bf Sentence example \\
  \hline
  Unnamed edges & Node A and Node B are connected.\\
  Named edges   & Edge B starts at Node A and goes to Node B.\\
  Edge distance & Edge B is 5 cm long.\\
  Edge distance & Node A and Node B have a distance of 5 cm.\\
  \hline\hline 
\end{tabular}
\end{center}

Nodes are specified implicitly by mentioned e.g. ``Node A'' somewhere
in the text.

Distances have a unit and magnitude. We only support ``cm'' as a unit.

Because we use triples as the basis of our application, we can
structure the sentences differently, while keeping the triple set the
same.

\begin{example}
  Extracting triples from the sentences

  \begin{quote}
    Node A and Node B are connected. Node A and Node B have a distance
    of 5 cm.
  \end{quote}

  and the sentence

  \begin{quote}
    Node A and Node B are connected with a distance of 5 cm.
  \end{quote}

  will result in the same set of triples.  
\end{example}

% Important: graph language we compile to needs to be declarative to keep the implementation simple

\section{Triple walks}

We define some helper goals that let us define a walk that follows a
chain of triples, and allows us to essentially pattern match on that chain.

\relation{(Xsubjecto->$\ t\ $&$\ clauses$)}

A subject walk succeeds if $t = (S_0,P_0,O_0)$ can
satisfy each of the $clauses$. $t$ is the triple we start with.

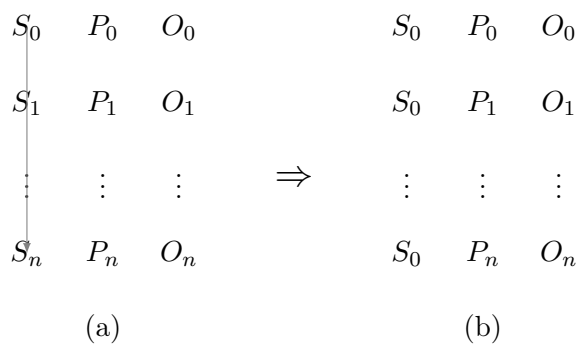
\begin{figure}[h!]
  \centering
  \begin{tikzpicture}[>=latex]
    \begin{scope}[local bounding box=scope1]
      \node (s1) {$S_0$};
      \node (p1) [right of=s1] {$P_0$};
      \node (o1) [right of=p1] {$O_0$};
      \node (s2) [below of=s1] {$S_1$};
      \node (p2) [right of=s2] {$P_1$};
      \node (o2) [right of=p2] {$O_1$};
      \node (dots1) [below of=s2] {\vdots};
      \node (dots2) [right of=dots1] {\vdots};
      \node (dots3) [right of=dots2] {\vdots};
      \node (sn) [below of=dots1] {$S_n$};
      \node (pn) [right of=sn] {$P_n$};
      \node (on) [right of=pn] {$O_n$};
      \path[draw,color=unused-edge,->]
      (s1.center) -- (s2.center) -- (dots1.center) -- (sn.center);
      \node (a)  [below of=pn] {(a)};
    \end{scope}

    \begin{scope}[xshift=5cm]
      \node (s1) {$S_0$};
      \node (p1) [right of=s1] {$P_0$};
      \node (o1) [right of=p1] {$O_0$};
      \node (s2) [below of=s1] {$S_0$};
      \node (p2) [right of=s2] {$P_1$};
      \node (o2) [right of=p2] {$O_1$};
      \node (dots1) [below of=s2] {\vdots};
      \node (dots2) [right of=dots1] {\vdots};
      \node (dots3) [right of=dots2] {\vdots};
      \node (sn) [below of=dots1] {$S_0$};
      \node (pn) [right of=sn] {$P_n$};
      \node (on) [right of=pn] {$O_n$};
      \node (b)  [below of=pn] {(b)};
    \end{scope}
    \node (arrow) [xshift=1.5cm,right of=scope1] {\Large $\Rightarrow$};
  \end{tikzpicture}
  \caption{Illustration of how a subject walk beginning with triple
    $t$, threads $t$'s subject through the whole triple chain.}
  \label{fig:subjecto}
\end{figure}

\begin{minipage}{\textwidth}
A clause can be a tuple with a predicate and object or a
3-tuple in which case the last element is a partial match of a word
map in the $S_0$'s word group.

We \emph{thread} the subject of $t$ through each of the clauses. This
is illustrated in \prettyref{fig:subjecto}.
\end{minipage}

\begin{example} We define the following subject walk to find an edge:  
\vspace*{0.5em}
\begin{tabular}{cccl}
  \sf $S_0$ & \sf :start & \sf $O_1$ & where a word map in $S_0$ must match \lstinline|{:lemma "Edge"}|\\
  \sf $S_0$ & \sf :at & \sf $O_1$ &\\
  \sf $S_0$ & \sf :go & \sf $O_2$ & \\
  \sf $S_0$ & \sf :to & \sf $O_2$ &
\end{tabular}
\vspace*{0.5em}

We use $O_1$ and $O_2$ in multiple clauses to make sure that the clauses only match the same
object group.

Looking at \prettyref{tab:reified-triples} we can find a triple for
which the walk succeeds: $t = (\textsf{edge-test-0}, \textsf{:start},
\textsf{node-start-0})$

\vspace*{0.5em}
\begin{tabular}{ccc}
  \sf edge-test-0 & \sf :start & \sf node-start-0\\
  \sf edge-test-0 & \sf :at & \sf node-start-0\\
  \sf edge-test-0 & \sf :go & \sf end-node-0\\
  \sf edge-test-0 & \sf :to & \sf end-node-0
\end{tabular}
\vspace*{0.5em}

We can run that walk with \Xsubjecto inside a \corelogic query:

\def\subjectospaces{\ \ \ \ \ \ \ \ \ \ \ \ \ \ \ }
\begin{lstlisting}
(Xsubjecto-> $t$
$\subjectospaces$[$\textsf{:start}$ $O_1$ {:lemma "Edge"}]
$\subjectospaces$[$\textsf{:at}$ $O_1$]
$\subjectospaces$[$\textsf{:go}$ $O_2$]
$\subjectospaces$[$\textsf{:to}$ $O_2$])
\end{lstlisting}

If the \Xsubjecto goal succeeds, $O_1$ will be bound to
\textsf{node-start-node-0} and $O_2$ to \textsf{end-node-0}.

We can then extract more information out of them, if necessary and
perform some additional validation.
\end{example}

\relation{(Xobjecto->$\ t\ $&$\ clauses$)}

\begin{figure}[h!]
  \centering
  \begin{tikzpicture}[>=latex]
    \begin{scope}[local bounding box=scope1]
      \node (s1) {$S_0$};
      \node (p1) [right of=s1] {$P_0$};
      \node (o1) [right of=p1] {$O_0$};
      \node (s2) [below of=s1] {$S_1$};
      \node (p2) [right of=s2] {$P_1$};
      \node (o2) [right of=p2] {$O_1$};
      \node (dots1) [below of=s2] {\vdots};
      \node (dots2) [right of=dots1] {\vdots};
      \node (dots3) [right of=dots2] {\vdots};
      \node (sn) [below of=dots1] {$S_n$};
      \node (pn) [right of=sn] {$P_n$};
      \node (on) [right of=pn] {$O_n$};
      \path[draw,color=unused-edge,->]
      (o1.center) -- (s2.center) -- (o2.center) -- (dots1.center) --
      (dots3.center) -- (sn.center) -- (on.center);
      \node (a)  [below of=pn] {(a)};
    \end{scope}

    \begin{scope}[xshift=5cm]
      \node (s1) {$S_0$};
      \node (p1) [right of=s1] {$P_0$};
      \node (o1) [right of=p1] {$O_0$};
      \node (s2) [below of=s1] {$O_0$};
      \node (p2) [right of=s2] {$P_1$};
      \node (o2) [right of=p2] {$O_1$};
      \node (dots1) [below of=s2] {\vdots};
      \node (dots2) [right of=dots1] {\vdots};
      \node (dots3) [right of=dots2] {\vdots};
      \node (sn) [below of=dots1] {$O_{n-1}$};
      \node (pn) [right of=sn] {$P_n$};
      \node (on) [right of=pn] {$O_n$};
      \node (b)  [below of=pn] {(b)};
    \end{scope}
    \node (arrow) [xshift=1.5cm,right of=scope1] {\Large $\Rightarrow$};
  \end{tikzpicture}
  \caption{Illustration of how an object walk beginning with triple
    $t$, threads the object of each consecutive triple through the triple chain.}
  \label{fig:objecto}
\end{figure}
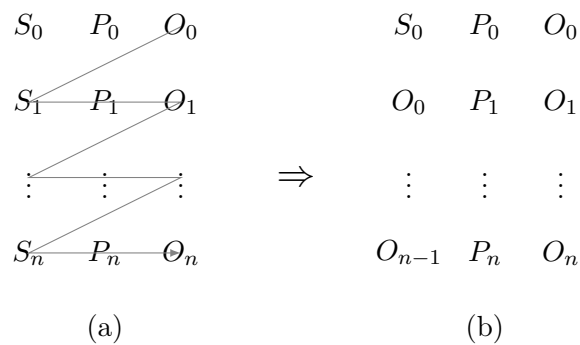

The counterpart to \Xsubjecto that matches on the triples' objects
first (see \prettyref{fig:objecto}).

\begin{example} We define an object walk to get the distance
  between the nodes of our edge. We assume that the previous subject
  walk succeeded with $t = (S_0, P_0, O_0)$.

\vspace*{0.5em}\begin{tabular}{cccl}
\sf $O_0$ & \sf :have & \sf $O_3$ & where a word map in \textsf{$O_3$} must match \lstinline|{:lemma "distance"}|\\
\sf $O_3$ & \sf :of & \sf $O_4$ & \\
\sf $O_4$ & \sf :be & \sf $O_5$ & where a word map in \textsf{$O_5$} must match \lstinline|{:tag "CD"}|\\
\end{tabular}\vspace*{0.5em}

The walk succeeds, if we start with the triple $t =
(\textsf{edge-test-0}, \textsf{:start},
\textsf{node-start-0})$ and follow the triples:

\vspace*{0.5em}
\begin{tabular}{ccc}
\sf node-start-0 & \sf :have & \sf distance-0\\
\sf distance-0 & \sf :of & \sf cm-2\\
\sf cm-2 & \sf :be & \sf num-10-0
\end{tabular}
\vspace*{0.5em}

We can run that walk with \Xobjecto inside a \corelogic query:

\def\objectospaces{\ \ \ \ \ \ \ \ \ \ \ \ \ \ }
\begin{lstlisting}
(Xobjecto-> $t$
$\objectospaces$[:have $O_3$ {:lemma "distance"}]
$\objectospaces$[:of $O_4$]
$\objectospaces$[:be $O_5$ {:tag "CD"}])
\end{lstlisting}

If the \Xobjecto goal succeeds and $O_3$, $O_4$, and $O_5$ are logic
variables, they will be bound to \textsf{distance-0}, \textsf{cm-2},
and \textsf{num-10-0} respectively.

We can then extract more information out of them, if necessary.
\end{example}

\newpage
\section{Implementation}

We implement several collection stages that search the triples for
relevant information. The general process follows these steps:
\begin{enumerate}
\item Collect node
\item Collect edges
\item Check found edges for inconsistencies
\item Check for singleton nodes
\end{enumerate}

Each stage might emit warnings about inconsistencies, which we need to
present to the user later.

The node and edge collecting stages make use of subject and object
walks to extract the information that we need to create a graph.

\begin{example}
If we combine the object walk with the subject walk in our previous
examples, we get an edge with the distance between its nodes:
\begin{itemize}
  \item The group $S_0 = \textsf{edge-test-0}$ contains the edge label
    (\emph{Test})
  \item The group $O_5 = \textsf{num-10-0}$ is the edge's length
  \item The group $O_4 = \textsf{cm-2}$ contains the edge length's unit
\end{itemize}

An edge's or node's label is extracted by using the word
that is immediately next to it in the original text.

Extracting the edge's length and unit is easy, because $O_5$ and $O_4$
are word groups with only a single word map, so we can simply use that
word map's lemma. We then check if the length is an integer and the
unit is one of the supported units (cm).
\end{example}

% If the subject walk succeeds first, $t$ will be bound to the triple with which the walk can
% succeed. $t$ is then used as the starting point of the object walk.

% \section{Summary}

The output of our example application is presented to the users as additional annotations:

\annotation{draw}{Chunk}{
  Shows the graph that was extracted from the chunk's text. The input
  text contains some ambiguities, which means the graph is not looking
  like we want it to look. The double edge between nodes ``Start'' and
  ``End'' looks especially suspicious. 
}{\includegraphics[width=311px]{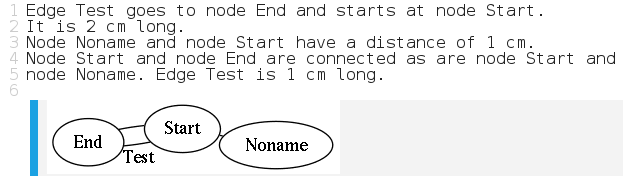}}

\annotation{draw-warnings \textrm{and} draw-warning-highlights}{Chunk}{
  The information collecting stages emit warnings about potential
  problem areas in our input text. We display some information
  about which sentences and words might be problematic and some
  suggestions about how to resolve the warnings.

  Here we can see that the edge ``Test'' is specified twice which
  explains the double edge we saw earlier.
}{\includegraphics[width=358px]{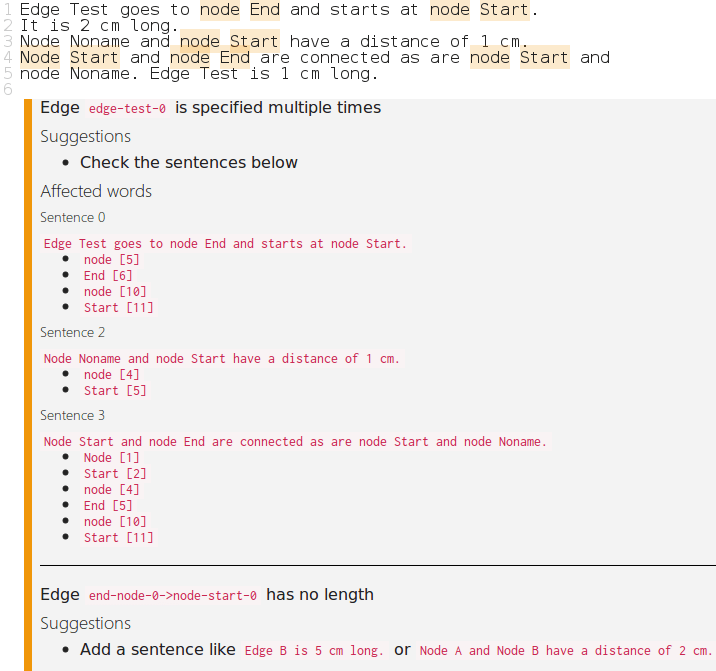}}

%\annotation{draw-warning-highlights}{Word}{
%  Highlights all words for which there are warnings.
%}{\includegraphics[width=301px]{annotations/draw-warning-highlights.png}}

% -*- TeX-master: "../master.tex"; ispell-local-dictionary: "en_US"; -*-

\chapter{Conclusion}
\label{cha:conclusion}

We have shown how we can extract information from a text in a
straightforward way, while only using the simple tools provided by
\corenlp, and what kind of problems result in integrating such a system
into a distributed development environment like \clide.

To achieve our initial goals, we have taken the following approach:
\begin{enumerate}
\item We provide an underlying NLP knowledge base (see
  \prettyref{sec:corenlp}) for an input text based on \corenlp's
  dependency parser and its coreference resolution system.
\item The reconciler (see \prettyref{sec:reconciler}) makes sure that
  text changes sent by \clide are 
  integrated into the knowledge base, and keeps the ontology and
  annotations up to date and in sync with the input text while
  making sure to only update them when really  necessary. Because
  coreference resolution is a slow process, and we need to rerun it
  after each text change, we split the input text into chunks and
  only operate on one chunk at a time.
\item Triple builders (see \prettyref{cha:triples}) are simple
  queries on the knowledge base that extract meaningful units of
  information from the text's semantic graphs in the form of triples
  $(subject,predicate,object)$.
\item The triples extracted from the queries are augmented by
  grouping their subjects and objects according to the coreference
  chain they belong to, yielding an ontology with unique classes
  (word groups).
  We provide an example for how to export an OWL ontology (see \prettyref{sec:ontology}), so that
  the ontology can be used by other tools.
\item We expose the underlying structure of sentences and texts to
  the user by providing annotations that visualize that structure in
  \clide (see \prettyref{sec:clide}).
\item In \prettyref{cha:usecase} we have shown how the ontology can be
  used to create graphs from a simple natural language
  specification and how using triples enables sentences in the input text
  with slightly different phrasing to still yield the same ontology
  and graph.
\end{enumerate}

There are some (solvable) caveats to our approach:
\begin{itemize}
\item Some triples we extract make no sense. This is a symptom of missing
  information and of the triple builders' simplistic nature. E.g.
  some triple builders like \textsf{nsubjpass-ccomp} are
  limited, because there is currently no way to determine a verb's
  transitivity.
\item At the moment we ignore all tenses and merge potentially different
  states of word groups in different time frames (as indicated e.g.
  by a triple's predicate's tense) into one ontology. Using
  a verb ontology in combination with the predicates' part-of-speech
  tags, we could split the ontology into separate ontologies, one
  for each time frame. We could then e.g. track the changes of a
  word group's attributes over time.
\item Currently the ontology we create has no concept hierarchy. We
  could integrate WordNet to group word groups that have the
  same concept behind them under one umbrella via e.g. \emph{is-a}
  relationships in the ontology.
\item At the moment the ontologies we extract are limited to one chunk
  of a text only. It should be possible to develop some heuristics
  that would allow us to merge the ontologies of two or more chunks
  together into one ontology.
\item \clidenlp is limited by \clide's current behavior to keep
  annotations static and to limit the assistants direct influence to the
  server-side only. A client side integration could enable some
  interactive aspects, like e.g. defining a new triple builder on the
  fly and using \clj's dynamic aspects to make it available to the system
  immediately.\footnote{This is possible already, but not exposed to
    the user in the UI.}
\end{itemize}

The tools we created while building the assistant are
general enough to be used outside of \clide and could be integrated
into other editing environments. Our NLP knowledge base with
information from \corenlp's semantic graphs, its coreference
resolution system, and the ontology we extracted from them, provides
easy access to information about a text.

Exposing \clj's and \clidenlp's dynamic natures in \clide's interface
and combining it with \clide's collaborative aspects can enable an
environment and framework where we can quickly and collaboratively
develop simple systems that have some albeit limited and domain
specific text understanding.

\appendix

\chapter{Part of Speech Tags}

The part of speech tags used by \corenlp are based on the
tags used by the Penn Treebank \citep{marcus1993}. There are however
some differences. \prettyref{tab:pos} shows an updated version of
the part of speech tag table in \citep[p. 317]{marcus1993} based on
experience with \corenlp. As such this table is most likely
incomplete, but enough to follow the examples in this report.

\begin{table}[h!]
  \caption{An incomplete list of part of speech tags used by \corenlp based on
    \citep[p. 317]{marcus1993}}
  \label{tab:pos}
  \begin{tabular}{lllll}
    \hline
    \sf CC & Coordinating conjunction & & TO & \emph{to} \\
    \sf CD & Cardinal number & & \sf UH & interjection\\
    \sf DT & Determiner & & \sf VB & Verb, base form\\
    \sf EX & Existential \emph{there} & & \sf VBD & Verb, past tense\\
    \sf FW & Foreign word & & \sf VBG & Verb, gerund/present
    participle\\
    \sf IN & Preposition/subordinating conjunction & & \sf VBN & Verb,
    past participle\\
    \sf JJ & Adjective & & \sf VBP & Verb, non-3rd ps. sing. present\\
    \sf JJR & Adjective, comparative  & & \sf VBZ & Verb, 3rd ps.
    sing. present\\
    \sf JJS & Adjective, superlative & & \sf WDT &
    \emph{wh}-determiner\\
    \sf LS & List item marker & & \sf WP & \emph{wh}-pronoun\\
    \sf MD & Modal & & \sf WP\$ & Possessive \emph{wh}-pronoun\\
    \sf NN & Noun, singular or mass & & \sf WRB & \emph{wh}-adverb\\
    \sf NNS & Noun, plural & & \sf RP & Participle \\
    \sf NNP & Proper noun, singular & & \sf \$ & Currency sign\\
    \sf NNPS & Proper noun, plural & & \sf . & Sentence-final
    punctuation \verb|! ? .|\\
    \sf PDT & Predeterminer & & \sf , & Comma\\
    \sf POS & Possessive ending & & \sf : & Colon, semi-colon\\
    \sf PRP & Possessive pronoun & & \sf -LRB- & Left bracket \verb|( [ {| character\\
    \sf PP\$ & Possessive pronoun & & \sf -RRB-  & Right bracket \verb|) ] }| character\\
    \sf RB & Adverb & & \sf \verb|``| & Left (double or single) quote\\
    \sf RBR & Adverb, comparative & & \sf \verb|''| & Right (double or
    single) quote\\
    \sf RBS & Adverb, superlative & & & \\
    \hline\hline
  \end{tabular}
\end{table}

\chapter{Installation notes}
\label{cha:install}

The CD that accompanies this report contains three ZIP files:

\begin{tabular}{ll}
  \bf Filename & \bf Contents\\
  \hline
  \tt report.pdf & A digital copy of this report\\
  \tt clide-nlp-src.zip & The source code for \clidenlp\\
  \tt clide-src.zip & The source code to a version of \clide that
  works correctly with \clidenlp\\
  \tt clide-nlp.zip & Contains an executable JAR of \clidenlp and
  startup scripts\\
  \hline\hline
\end{tabular}

\clidenlp requires Java 7 and \clide works best with a WebKit-based
browser like e.g. Chrome. You need approx. 3 GiB of RAM to
successfully run \clidenlp. To execute it:
\begin{itemize}
  \item Extract \texttt{clide-nlp.zip} and run
    \texttt{run.sh} on Linux/FreeBSD or \texttt{run.bat} on Windows.
  \item Wait a minute or two.
  \item A launcher window will pop up that informs you about the
    startup process.
  \item After the system is ready, \clidenlp should inform you that it is ready at
    \url{http://localhost:14000}
  \item Open the URL and log in with user \texttt{clide-nlp} and
    password \texttt{clide-nlp}.
  \item Open the \texttt{clide-nlp/Example} project and
    look at \texttt{00-README.txt} for further help.
  \item And most importantly, try editing one of the files!
\end{itemize}

\cleardoublepage
\phantomsection
\addcontentsline{toc}{chapter}{List of Figures}
\listoffigures

\bibliography{literature}

% \cleardoublepage
% \phantomsection
% \addcontentsline{toc}{chapter}{ToDo}
% \listoftodos

\end{document}